%% file: paper.tex
\documentclass[]{opendatalab}
\usepackage{caption}
\usepackage{xcolor}
\usepackage{longtable}
\usepackage{listings}
\usepackage[numbers]{natbib}
\usepackage{amssymb}
\usepackage{tabularx}
\usepackage{float}
\usepackage{graphicx}
\usepackage{makecell}
\usepackage[table]{xcolor}
\usepackage{cleveref}
\usepackage{array}
\usepackage[misc]{ifsym}
\usepackage{tcolorbox}
\usepackage{enumitem}
\usepackage{listings}
\tcbuselibrary{skins, breakable}
\usepackage{capt-of} 
\newtcolorbox{promptbox}[1]{
    colback=gray!5,
    colframe=black!75,
    left=2mm, right=2mm, top=2mm, bottom=2mm,
    fonttitle=\bfseries,
    title=#1,
    sharp corners,
    boxrule=0.5pt
}
\newcommand{\redapo}{Re$^3$-DAPO}

\definecolor{codegreen}{rgb}{0,0.6,0}
\definecolor{codegray}{rgb}{0.5,0.5,0.5}
\definecolor{codepurple}{rgb}{0.58,0,0.82}
\definecolor{backcolour}{rgb}{0.95,0.95,0.92}
\definecolor{promptcolor}{HTML}{D1D0F2}
\definecolor{promptcolorheader}{HTML}{bdbcec}
\makeatletter
\newcommand\andauthor{%
  \g@addto@macro\authorlist{\par\vspace{1mm}}
}
\newcommand\nlauthor[2][]{%
  \addtolist[#1]{#2}{\authorlist}{\authorformat}{}
}
\makeatother

\RequirePackage{xspace}
\makeatletter
\DeclareRobustCommand\onedot{\futurelet\@let@token\@onedot}
\def\@onedot{\ifx\@let@token.\else.\null\fi\xspace}

\makeatother

\definecolor{codegreen}{rgb}{0,0.6,0}
\definecolor{codegray}{rgb}{0.5,0.5,0.5}
\definecolor{codepurple}{rgb}{0.58,0,0.82}
\definecolor{backcolour}{rgb}{0.95,0.95,0.92}
\definecolor{promptcolor}{HTML}{D1D0F2}
\definecolor{promptcolorheader}{HTML}{bdbcec}
\lstdefinestyle{mystyle}{
    backgroundcolor=\color{backcolour},   
    commentstyle=\color{codegreen},
    keywordstyle=\color{magenta},
    numberstyle=\tiny\color{codegray},
    stringstyle=\color{codepurple},
    basicstyle=\ttfamily\footnotesize,
    breakatwhitespace=false,         
    breaklines=true,                 
    captionpos=b,                    
    keepspaces=true,                 
    numbers=left,                    
    numbersep=5pt,                  
    showspaces=false,                
    showstringspaces=false,
    showtabs=false,                  
    tabsize=2
}

\title{Molecular Identifier Visual Prompt and \\Verifiable Reinforcement Learning for Chemical Reaction Diagram Parsing}
\vspace{10pt}

\author[1,2*]{Jiahe Song}
\author[3,2*]{Chuang Wang}
\author[2*]{Yinfan Wang}
\author[5*]{Hao Zheng}
\author[3,2*]{Rui Nie}
\author[4,2*]{Bowen Jiang}
\andauthor

\nlauthor[2]{Xingjian Wei}
\author[2]{Junyuan Gao}
\author[2]{Yubin Wang}
\author[2]{Bin Wang}
\author[2]{Lijun Wu}
\andauthor
\nlauthor[2\ \textrm{\Letter}]{Jiang Wu}
\author[3\ \textrm{\Letter}]{Qian Yu}
\author[2\ \textrm{\Letter}]{Conghui He}

\affiliation[1]{Shanghai Jiao Tong University}
\affiliation[2]{Shanghai AI Laboratory}
\affiliation[3]{Beihang University}
\affiliation[4]{Peking University}
\affiliation[5]{South China Normal University}
\abstract{
Reaction diagram parsing (RxnDP) is critical for extracting chemical synthesis information from literature. Although recent Vision-Language Models (VLMs) have emerged as a promising paradigm to automate this complex visual reasoning task, their application is fundamentally bottlenecked by the inability to align visual chemical entities with pre-trained knowledge, alongside the inherent discrepancy between token-level training and reaction-level evaluation. To address these dual challenges, this work enhances VLM-based RxnDP from two complementary perspectives: prompting representation and learning paradigms. First, we propose \textbf{Identifier as Visual Prompting (IdtVP)}, which leverages naturally occurring molecule identifiers (e.g., bold numerals like \textbf{1a}) to activate the chemical knowledge acquired during VLM pre-training. IdtVP enables powerful zero-shot and out-of-distribution capabilities, outperforming existing prompting strategies. Second, to further optimize performance within fine-tuning paradigms, we introduce \textbf{\redapo}, a reinforcement learning algorithm that leverages verifiable rewards to directly optimize reaction-level metrics, thereby achieving consistent gains over standard supervised fine-tuning. Additionally, we release the \textbf{ScannedRxn} benchmark, comprising scanned historical reaction diagrams with real-world artifacts, to rigorously assess model robustness and out-of-distribution ability. Our contributions advance the accuracy and generalization of VLM-based reaction diagram parsing. We will release data, models, and code on GitHub.
}

\date{\today}
\equalcontribution{Jiahe Song, Chuang Wang, Yinfan Wang, Hao Zheng, Rui Nie, Bowen Jiang}
\projectlead{Jiang Wu}

\correspondence{Conghui He, \email{heconghui@pjlab.org.cn}, Jiang Wu, \email{wujiang@pjlab.org.cn}, Qian Yu, \email{qianyu@buaa.edu.cn}} 

\usepackage[T1]{fontenc}
\usepackage[utf8]{inputenc}
\begin{document}

\maketitle

\input{sec/1_intro}
\input{sec/3_pilot}
\input{sec/4_method}
\input{sec/5_exp}

\input{sec/6_conclusion}


\clearpage
\newpage
\bibliographystyle{plainnat}
\setcitestyle{numbers}

\clearpage
\newpage
\beginappendix

\input{sec/sup}

\bibliography{paper}
\end{document}

%% file: sec/1_intro.tex
\section{Introduction}
\label{sec:intro}

High-quality chemical reaction datasets form the cornerstone of AI for Science research~\cite{ding2025survey}, driving the development of automated reaction information extraction from chemical literature~\cite{schneider2016big,fan2024openchemie,zhong2023reaction}. Although existing pipelines heavily rely on text mining, complete synthesis routes and complex mechanisms are primarily communicated through graphical diagrams. As a result, Reaction Diagram Parsing (RxnDP) has emerged as an indispensable task to capture this visually encoded knowledge. Because these diagrams encapsulate rich topological and stereochemical details often omitted in plain text, RxnDP is crucial for unlocking the massive, high-quality data trapped in unstructured bitmaps—a prerequisite for training comprehensive chemical foundation models. Furthermore, visual parsing provides unique structural insights that enable cross-validation with text-based data (as detailed in \cref{sec:crossmodel}), ultimately facilitating the construction of highly accurate chemical databases.

While traditional deep learning methods have been widely applied to RxnDP, Vision-Language Models (VLMs)~\cite{gemini3pro2026, bai2025qwen2, hurst2024gpt, li2025uni} have demonstrated immense potential. Benefiting from breakthroughs in visual and multimodal reasoning, VLMs surpass traditional deep learning approaches~\cite{guo2021automated,zhu2025tamer,zhong2023reactie,qian2023rxnscribe, wilary2023reactiondataextractor}. This superiority stems from two advancements: first, the extensive world knowledge acquired through large-scale interleaved image-text pre-training~\cite{li2024omnicorpus, zhu2023multimodal, laurenccon2023obelics}; second, the significant enhancement of cross-domain generalization achieved through Reinforcement Learning~\cite{guo2025deepseek,yu2025perception, yu2025dapo, hu2025open,huang2025vision,shen2025vlm, wen2025reinforcement} from Verifiable Rewards (RLVR)~\cite{wang2025rlver,shao2024deepseekmath,gao2410designing,lambert2024tulu} during post-training, moving beyond standard fine-tuning.

Despite these significant advancements in pre-training and RLVR, their specific application to RxnDP remains under-explored. This study aims to address \textbf{two core questions}: (1) \textit{How can we better leverage the pre-trained core capabilities of VLMs for chemical diagram understanding?} and (2) \textit{How can we optimize RLVR to improve the performance and generalization of RxnDP?}

\begin{figure*}[t]
    \centering
    \includegraphics[width=\textwidth]{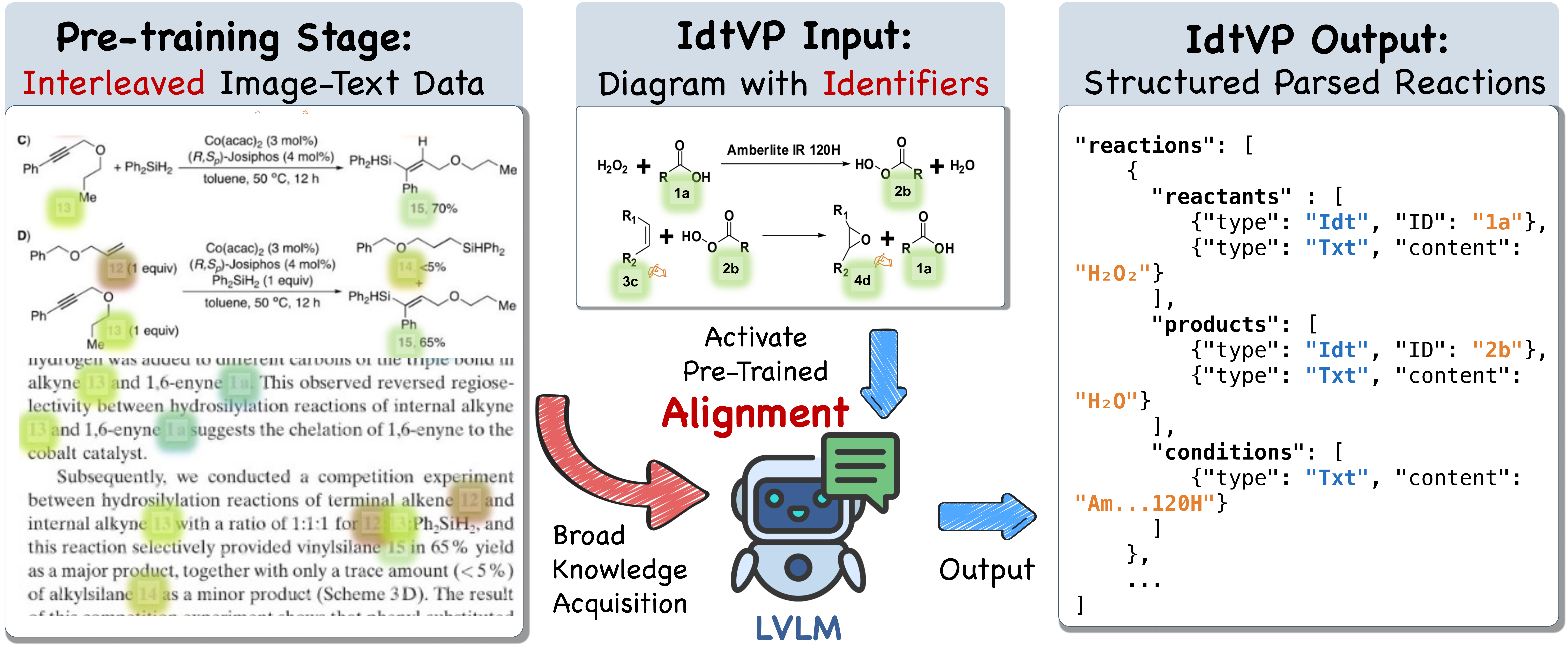}
    \caption{Overview of the \textbf{Id}en\textbf{t}ifier as \textbf{V}isual \textbf{P}rompting (\textbf{IdtVP}) strategy. \textbf{Left:} VLMs learn visual-text alignment from identifier-rich interleaved literature during pre-training. \textbf{Middle:} IdtVP inputs diagrams annotated with molecule identifiers to activate this pre-trained knowledge. \textbf{Right:} The model outputs structured reactions, directly using these identifiers as ``handles'' to concisely represent complex molecules.}
    \label{fig:IdtVP_teaser}
\end{figure*}

\textbf{Regarding the first challenge}, prior work~\cite{song2025rxncaption} indicated that the conventional \textbf{BROS} (\textbf{B}box and \textbf{R}ole in \textbf{O}ne \textbf{S}tep) approach is suboptimal for VLMs. To address this, they proposed \textbf{BIVP} (\textbf{B}ox \textbf{I}ndexing as \textbf{V}isual \textbf{P}rompts), which detects molecules and overlays indexed bounding boxes onto the image as visual prompt. However, this practice introduces a distribution shift from the raw images used in VLM pre-training. We observe that chemical literature frequently uses bold numerals as \textit{molecule identifiers} to annotate reaction diagrams, and the main text commonly employs similar shorthand. Because VLM pre-training corpora contain vast amounts of such interleaved image-text data, models are inherently accustomed to describing chemical reactions using these reference numbers. Capitalizing on this, we propose \textbf{IdtVP} (\textbf{Id}entifier as \textbf{V}isual \textbf{P}rompting), which utilizes the inherent identifiers in reaction diagrams as descriptive anchors. Experiments demonstrate that IdtVP significantly outperforms BROS and BIVP in both zero-shot and fine-tuning settings. For instance, on Gemini 3.0 Pro~\cite{gemini3pro2026}, our approach achieves state-of-the-art zero-shot Soft Match F1 scores, proving that IdtVP unlocks the chemical knowledge embedded in VLM pre-training. Furthermore, models supervised fine-tuned using the IdtVP strategy surpass those using BIVP, while also exhibiting stronger generalization.

\textbf{Regarding the second challenge}, current RxnDP fine-tuning relies entirely on Supervised Fine-Tuning (SFT)~\cite{qian2023rxnscribe,RxnIm, song2025rxncaption}, which leads to two key limitations. First, \textbf{token-level objectives misalign with final evaluation metrics}. SFT provides token-wise supervision signals, whereas the task requires structured JSON outputs. Second, \textbf{the forced serialization of set-based tasks introduces noise}. Since RxnDP is inherently order-agnostic, SFT compels models to learn dataset-specific sequence orderings, consequently leading to overfitting, as shown in \cref{sec:re3_dapo}.

To resolve this issue, we introduce the \textbf{\redapo}~algorithm, employing verifiable reinforcement learning. Its core innovation lies in parsing the model-generated JSON outputs during reward computation and directly translating evaluation metrics into dense reward signals. This elevates the optimization objective from the token level to the reaction level, simultaneously eliminating serialized supervision signals. Experiments reveal that this method achieves significant performance gains over SFT baselines and proves effective under both BIVP and IdtVP strategies.

To evaluate model generalization, we introduce \textbf{ScannedRxn}, a unique benchmark dataset sourced from printed literature spanning the 1950s to the 1990s. It features authentic scanning artifacts and era-specific typography, diverging significantly from modern digital standards. This fills a critical gap present in existing natively-digital datasets, providing a rigorous testbed for assessing model robustness during the digitization of historical literature.

In summary, the main contributions of this paper are:
\begin{itemize}
    \item \textbf{IdtVP Strategy:} We propose the ``Identifier as Visual Prompt(IdtVP)'' strategy, aligning with VLM pre-training patterns to activate built-in chemical knowledge and achieve robust zero-shot generalization.
    \item \textbf{\redapo ~Algorithm:} We introduce the \redapo ~algorithm, combining RLVR with mixed reward signals to address the misalignment of token-level objectives and serialization noise in set prediction tasks. Comprehensive experiments validate its significant performance improvements.
    \item \textbf{ScannedRxn Benchmark:} We release ScannedRxn, a highly challenging benchmark dataset containing printed literature from the 1950s to 1990s alongside authentic scanning artifacts, offering a rigorous testbed for evaluating the robustness of legacy chemical knowledge digitization.
\end{itemize}

%% file: sec/3_pilot.tex
\section{Identifier as Visual Prompt strategy}
\label{pilot_study}

\subsection{Reaction Diagram Parsing Task}
\label{sec:rxndp_task}

We formulate the Chemical Reaction Diagram Parsing (RxnDP) task as follows. Given an input diagram \( I \), the RxnDP model \( \mathcal{M} \) extracts a structured representation of all contained reactions, denoted as the set \( \mathcal{R} = \{r_k\} \). This relationship is expressed as:
\(
\mathcal{R} = \mathcal{M}(I).
\)
Each individual reaction \( r_k \in \mathcal{R} \) is defined as a triple encompassing three key roles:
\(
r_k = \left( \mathcal{S}_{\text{reactants}},\; \mathcal{S}_{\text{conditions}},\; \mathcal{S}_{\text{products}} \right).
\)
For every role \( \mathcal{S}_{\text{role}} \) (where \(\text{role} \in \{\text{reactants, conditions, products}\}\)), its contents form a set of components: \( \mathcal{S}_{\text{role}} = \{ e_m \} \). These components belong to one of two modalities:
\textbf{1. Molecular Components:} Represented by the coordinates of the bounding box surrounding the corresponding molecular structure in the image \( I \).
\textbf{2. Textual Components:} Represented directly by their string-based textual description.

We adopt the reaction-level evaluation protocol from RxnCaption~\cite{song2025rxncaption}, reporting Precision, Recall, and F1 scores. A predicted reaction $R^{\text{pred}}$ is a True Positive (TP) if it matches a ground truth $R^{\text{gt}}$ based on the following metrics:
\textbf{Soft Match}: Ignores textual content and merges conditions into reactants. A match requires an Intersection over Union (IoU) $> 0.5$ for all molecular bounding boxes.
\textbf{Hybrid Match}: Tailored to output formats. For \textit{BROS}: Adheres to the RxnScribe Hard Match standard, requiring IoU $> 0.5$ for all components (reactants, products, and conditions). For \textit{BIVP/IdtVP}: Decouples structure and text. Molecules require IoU $> 0.5$, while textual components are validated via string similarity, requiring a normalized edit distance $\leq 0.2$ against the corresponding ground truth text.

\subsection{Our Insights and IdtVP Strategy}

For RxnDP, prior methods use \textbf{BROS} (\textbf{B}box and \textbf{R}ole in \textbf{O}ne \textbf{S}tep) to directly regress molecular coordinates. RxnCaption~\cite{song2025rxncaption} argued this is ill-suited for VLMs and proposed \textbf{BIVP} (\textbf{B}ox \textbf{I}ndexing as \textbf{V}isual \textbf{P}rompts): detect molecules via MolYOLO, then overlay indexed boxes for the model to predict indices. BIVP achieved strong performance (MolYOLO >98\% precision/recall) with faster inference, providing crucial inspiration. However, pre-drawing boxes creates a distribution shift from VLM pretraining's raw-image paradigm, inhibiting full activation of pretrained chemical knowledge.

We observe that, as illustrated in~\cref{fig:IdtVP_teaser}, chemical literature (particularly reaction diagrams) frequently annotates molecular structures with boldface numerals as identifiers (termed ``molecule identifiers''). Such identifiers are not only ubiquitous within diagrams but also pervasive in accompanying paper text. Indeed, when describing reactions, chemical papers typically avoid cumbersome IUPAC names or SMILES strings, instead preferring concise identifiers to form expressions like ``Using reactants \textbf{1} and \textbf{2} as starting materials, under catalysis by compound \textbf{3}, to afford product \textbf{4}.'' In VLM pretraining corpora, such identifier-rich text often appears paired or interleaved with reaction diagrams, naturally conditioning models to adopt these identifiers as the preferred ``language'' for describing chemical reactions.

Building upon this insight, we propose the \textbf{Id}en\textbf{t}ifier as \textbf{V}isual \textbf{P}rompting  (\textbf{IdtVP}) strategy for RxnDP, with the workflow depicted in~\cref{fig:pipeline}:
(1) For molecules lacking identifiers in the chemical reaction diagram, we manually draw a unique, non-conflicting numerical identifier at an appropriate position adjacent to each molecule, ensuring every molecular structure is assigned a distinct identifier.
(2) The model is then prompted to output chemical reactions using these molecular identifiers directly as ``handles'' to refer to corresponding molecules, as shown in~\cref{fig:IdtVP_teaser}.

\subsection{Preliminary Verification of IdtVP Strategy}

\subsubsection{Zero-Shot Evaluation}

\input{Tables/zero_shot}

To validate the hypothesis that identifiers serve as visual prompt to activate the inherent chemical domain priors within VLMs, we conducted a preliminary zero-shot experiment comparing three strategies: BROS, BIVP, and IdtVP using representative models like Gemini 3 Pro~\cite{gemini3pro2026}. As shown in~\cref{tab:zero_shot_comparison}, the BROS strategy (raw image input) failed to generate valid graph structures. While the BIVP strategy (bounding box prompting) provided spatial guidance, its performance remained limited, achieving only 45.27\% Hybrid Match F1 on the RxnScribe-test dataset.

In contrast, the IdtVP strategy demonstrated significant superiority, boosting Gemini 3 Pro's performance to 71.70\% Hybrid Match F1 (and 88.11\% Soft Match F1) on RxnScribe-test, and maintaining a leading position on the more challenging RxnCaption-15k-test (78.23\% Soft Match F1). These results strongly suggest that explicitly overlaid identifiers effectively bridge the semantic gap between visual features and reaction logic, thereby unlocking the model's latent domain knowledge. Furthermore, the performance drop observed when applying IdtVP to a smaller parameter model (Qwen2.5-VL-7B~\cite{bai2025qwen2}, Intern-S1~\cite{bai2025intern}) indicates that high-precision zero-shot parsing requires a synergy of effective visual guidance and robust reasoning capabilities in the foundation model.


\subsubsection{Visual Attention Analysis}

\begin{figure}[h]
    \centering
    \includegraphics[width=\linewidth]{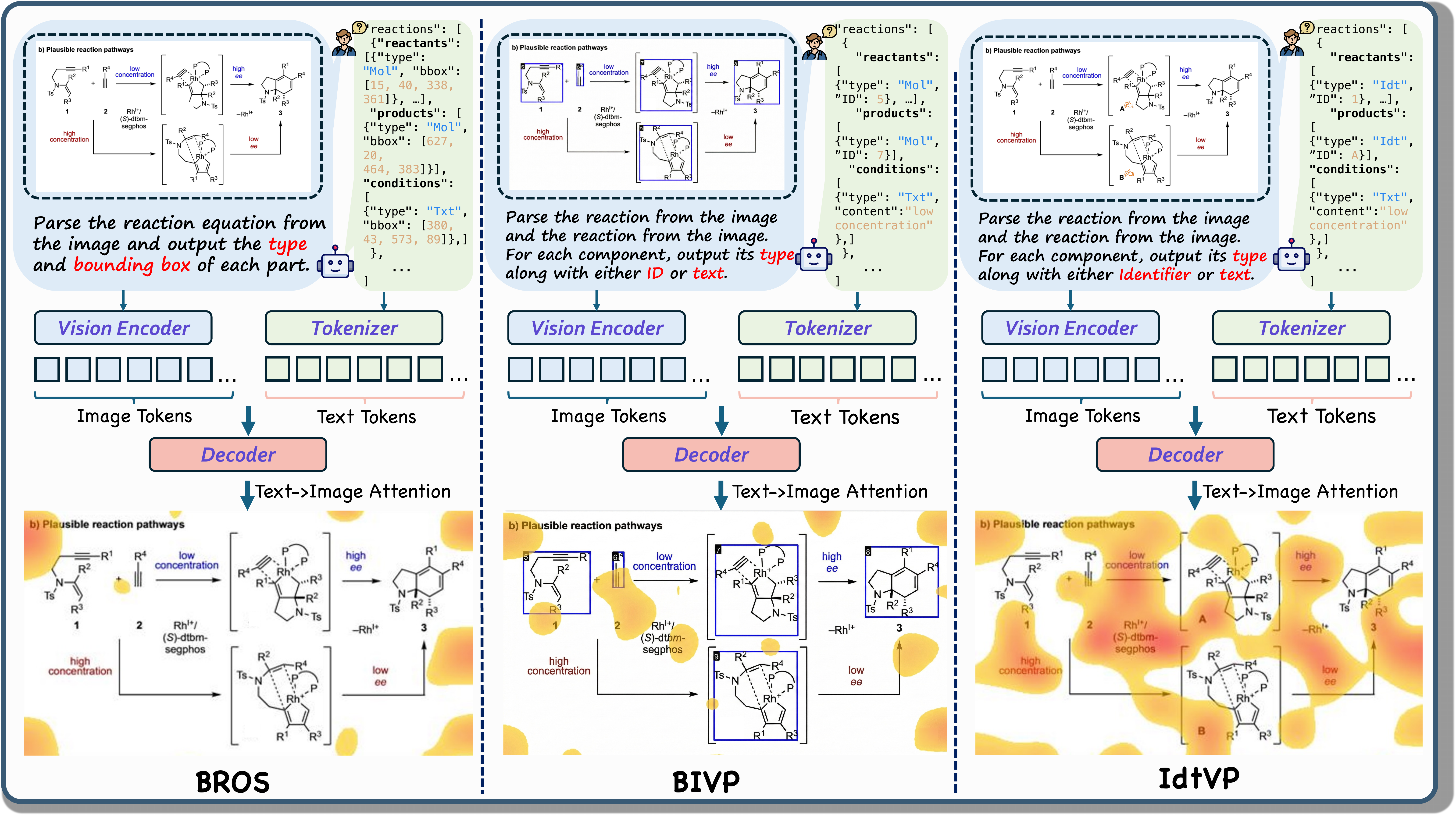}
    \caption{\textbf{Attention heatmaps of different prompting strategies.} The top panels illustrate the input images, textual prompts, and corresponding output formats for each strategy. The bottom panels visualize the Text-to-Image attention heatmaps, computed by averaging the attention weights across all heads per layer, followed by average pooling across all layers. Notably, IdtVP yields far more precise and comprehensive visual grounding on molecules and text compared to BROS and BIVP.}
    \label{fig:attention}
\end{figure}
To further validate our hypothesis, we adopt Layer-wise Attention Pooling~\cite{oh2022don} to visualize the cross-modal attention heatmaps of Qwen2.5-VL-7B~\cite{bai2025qwen2} during the response generation stage, which reveal the visual features the model relies on when producing molecular or establishing referential relationships.

As illustrated in~\cref{fig:attention}, the BROS strategy exhibits dispersed attention distributions, suggesting that the model struggles to establish strong semantic connections between the coordinate regression task and concrete chemical structures. The BIVP approach introduces explicit visual prompt in the form of bounding boxes to guide attention toward target regions. Although this leads to more localized attention, the activations are frequently concentrated around the bounding box boundaries rather than the corresponding referential index. In contrast, the IdtVP strategy produces more concentrated and precise attention patterns, with activations consistently aligned with the identifiers and their molecular structures. This observation provides evidence for our hypothesis that VLMs may already encode latent representations that associate ``identifiers'' (e.g., 1a, 2b) with molecular entities from their pretraining. By constructing visual context that is more consistent with such representations, IdtVP appears to facilitate the activation of this latent knowledge without additional fine-tuning, leading to improved domain behavior compared to alternative strategies.

%% file: Tables/zero_shot.tex
\begin{table}[t]
\centering
\caption{\textbf{Comparison of different visual prompting strategies under zero-shot setting.} H-F1 and S-F1 denote Hybrid-F1 and Soft-F1 scores, respectively (\%).}
\label{tab:zero_shot_comparison}
\setlength{\tabcolsep}{4pt}

\begin{tabular}{lcccc}
\toprule
\multirow{2}{*}{\textbf{Strategy}} & \multicolumn{2}{c}{\textbf{RxnScribe-test}} & \multicolumn{2}{c}{\textbf{RxnCaption-15k-test}} \\ 
\cmidrule(lr){2-3} \cmidrule(lr){4-5}
 & \textbf{H-F1} & \textbf{S-F1} & \textbf{H-F1} & \textbf{S-F1} \\
\midrule
Gemini 3 Pro + BROS  & 3.90  & 71.20 & 4.92  & 49.80 \\
Gemini 3 Pro + BIVP  & 45.27 & 81.32 & 47.40 & 73.40 \\
Gemini 3 Pro + IdtVP & \textbf{71.70} & \textbf{88.11} & \textbf{58.30} & \textbf{78.23} \\
\midrule
Qwen2.5-VL-7B + IdtVP & 16.69 & 39.00 & 4.88  & 19.16 \\
Intern-S1 + IdtVP     & 44.71 & 60.84 & 22.69 & 36.45 \\
\bottomrule
\end{tabular}
\vspace{-1em}
\end{table}

%% file: sec/4_method.tex
\section{Method}

\begin{figure*}[t]
    \centering
    \includegraphics[width=\textwidth]{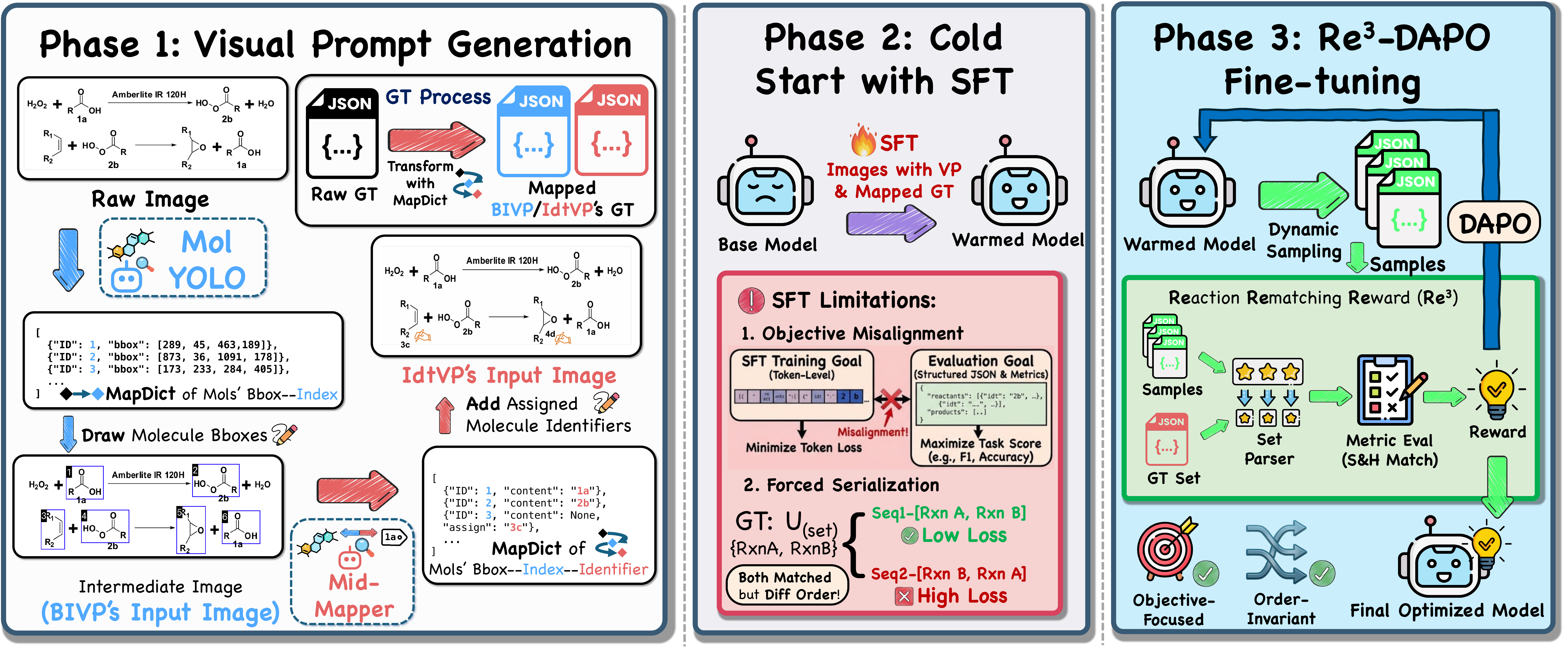}
    \caption{\textbf{Overview of the proposed framework.} Phase \textbf{1} constructs IdtVP data effectively supporting both \textbf{zero-shot inference} and \textbf{model training}. Phases \textbf{2-3} introduce a generalized optimization paradigm (SFT followed by \redapo) that is transferable to other prompting variants (e.g., BIVP).}
    \label{fig:pipeline}
    \vspace{-5pt}
\end{figure*}

\subsection{Overall Framework}

As illustrated in~\cref{fig:pipeline}, the proposed framework parses chemical reaction diagrams via a progressive three-stage pipeline: Visual Prompt Generation, Cold Start with SFT and \redapo~Fine-tuning.

In Phase \textbf{1}, we utilize our Mid-Mapper (\cref{sec:I-Det}) to construct the IdtVP by explicitly mapping identifiers to molecular bounding boxes. This format supports zero-shot inference and serves as a high-quality foundation for model training.

Phase \textbf{2} establishes the model's initial capabilities via SFT. While this serves as a necessary cold start, it inevitably inherits the optimization bottlenecks discussed in \cref{sec:intro} (i.e., objective misalignment and permutation sensitivity). To overcome these limitations, Phase \textbf{3} introduces the \redapo~fine-tuning strategy (\cref{sec:re3_dapo}). By shifting the learning objective from \textbf{next-token prediction} to \textbf{set-level matching}, we align the optimization directly with the evaluation metrics. Notably, this training curriculum (Phase 2 $\rightarrow$ Phase 3) constitutes a generalized optimization paradigm, applicable not only to IdtVP but also transferable to other prompting variants (e.g., BIVP).

\subsection{Mid-Mapper: Molecule-Identifier Mapper}
\label{sec:I-Det}
To establish a semantic association between chemical identifiers and molecular structures in reaction diagrams, we propose \textbf{M}olecule–\textbf{Id}entifier \textbf{Mapper} (\textbf{Mid-Mapper}). Leveraging the precise molecular localization provided by MolYOLO, the Mid-Mapper enables accurate alignment between each detected molecular structure and its corresponding chemical identifier.

Specifically, we distilled Gemini 3 Pro to construct high-quality training data that maps visual localization signals to semantic identifiers. For molecules that lack explicit identifiers in the original reaction diagrams, we required Gemini to assign context-consistent virtual identifiers based on reaction mechanism, ensuring that every molecular structure localized by MolYOLO has a unique and non-overlapping identifier. Building upon this dataset, we performed full-parameter fine-tuning of Qwen2.5-VL-3B to learn robust molecule–identifier associations. At inference time, the predicted virtual identifiers are plotted back into the original diagrams, thereby providing complete and semantically aligned visual annotations. This design ensures that the downstream IdtVP framework can perform efficient reasoning based on comprehensive and well-grounded visual features. The details of Mid-Mapper are provided in Appendix B.

\subsection{\redapo ~Training}
\label{sec:re3_dapo}

Standard Supervised Fine-Tuning (SFT) relies on token-level teacher-forcing, making it highly sensitive to the serialization order of target data. This is particularly problematic for the RxnDP task, where chemical reaction diagrams depict multiple independent reactions and entities that are inherently permutation-invariant. To quantify this limitation, we evaluated fully converged SFT models (BIVP and IdtVP) on the training set (see \cref{sec:data_construct} for details on dataset construction), explicitly isolating samples where the models predicted chemically perfect structures. As shown in~\cref{tab:sft_inconsistency_breakdown}, even on these semantically accurate samples, both models incurred considerable cross-entropy loss and generated non-zero gradients solely due to formatting mismatches with the ground truth.

\input{Tables/sft_error}

This confirms that rigid teacher-forcing compels the network to memorize arbitrary data formats rather than genuine chemical semantics. To address this disconnect between token-level optimization and order-agnostic, reaction-level accuracy, we propose the \redapo~algorithm. \redapo~leverages verifiable reinforcement learning to compute rewards based purely on the structured chemical entities, effectively bypassing the noise of arbitrary serialization.

Specifically, we define a composite, permutation-invariant reward function $R(y)$ that translates discrete evaluation metrics into a dense signal. As introduced in \cref{sec:rxndp_task}, it averages Hybrid Match ($R_{\text{hybrid}}$) and Soft Match ($R_{\text{soft}}$) to prioritize molecular accuracy over rigid role categorization:

\begin{equation}
R(y) = \frac{1}{2} \left( R_{\text{hybrid}}(y, y_{gt}) + R_{\text{soft}}(y, y_{gt}) \right)
\end{equation}

We employ DAPO to directly maximize this expected reward. By elevating the optimization from the token level to the reaction level, \redapo~allows the model to explore valid JSON serializations anchored solely by an order-invariant semantic objective, eliminating the need for complex actor-critic architectures.

\subsection{ScannedRxn Dataset}
\label{sec:ScannedRxn}

Existing chemical reaction parsing benchmarks, such as RxnCaption-15k-test~\cite{song2025rxncaption} and RxnScribe-test~\cite{qian2023rxnscribe}, are primarily derived from contemporary literature and lack the printed data styles common in the 20th century. Consequently, they fail to fully evaluate model generalization on historical archives, which is crucial for comprehensively extracting chemical knowledge and preserving valuable scientific discoveries. To bridge this gap, we introduce the ScannedRxn dataset. Collected from publications spanning the 1950s to the 1990s, this dataset comprises 200 reaction diagrams evenly distributed across four types: single, multi-line, tree, and cyclic (50 samples each). It effectively captures the distribution of traditional printed styles, including scanning noise, typewriter fonts, and legacy layout conventions. Comprehensive construction details and illustrative examples of ScannedRxn are provided in Appendix C.

%% file: Tables/sft_error.tex
\begin{table}[ht]
\centering
\caption{\textbf{Analysis of sequential inconsistency in SFT models.} Image-level statistics exclude single-reaction samples.}
\label{tab:sft_inconsistency_breakdown}
\setlength{\tabcolsep}{5pt} 
\renewcommand{\arraystretch}{0.9} 
\begin{tabular}{lcccccc} 
\toprule
\textbf{Model} & \multicolumn{3}{c}{\textbf{Image-level}} & \multicolumn{3}{c}{\textbf{Reaction-level}} \\
\cmidrule(lr){2-4} \cmidrule(lr){5-7}
& Total & Errors & Rate (\%) & Total & Errors & Rate (\%) \\
\midrule
BIVP  & 3,882 & 500 & \textbf{12.88} & 20,992 & 3,131 & \textbf{14.92} \\ 
IdtVP & 5,833 & 272 & \textbf{4.66}  & 31,300 & 2,538 & \textbf{8.11}  \\
\bottomrule
\end{tabular}
\vspace{-1em}
\end{table}

%% file: sec/5_exp.tex
\section{Experiment}
\input{Tables/exp}

\subsection{Experiment Setup}

\subsubsection{Dataset Construction} 
\label{sec:data_construct}
To construct the training dataset for IdtVP (denoted as RxnID), we built upon the official open-source RxnCaption-15k dataset. Specifically, we utilized the trained Mid-Mapper to detect existing identifiers within these reaction diagrams. For molecules lacking explicit identifiers, we assigned new ones and employed automated tools to directly draw them onto the original images. Both the resulting dataset configuration and the adopted data augmentation strategies strictly follow the RxnCaption~\cite{song2025rxncaption} setting. Meanwhile, the SFT and RL (\redapo) training of BIVP were directly performed on the unmodified RxnCaption-15k dataset.

\subsubsection{Mid-Mapper} Mid-Mapper was developed by performing full-parameter SFT on Qwen2.5-VL-3B via the ms-swift framework. Training was conducted on two NVIDIA A100 GPUs for 1 epoch with a learning rate of $1 \times 10^{-5}$, a 5\% warmup ratio, and a per-device batch size of 2.

\subsubsection{RxnDP Model}

The training of RxnCaption and RxnID consisted of two stages: 

\begin{itemize}
    \item \textbf{Cold-Start SFT}: To ensure schema adherence, we performed an SFT cold-start using approximately 10\% of the training data, deliberately selecting the checkpoint where the loss curve began to plateau. This brief SFT prevents model memorization while providing a robust initialization for JSON schema constraints. We fine-tuned the model using ms-swift on 8 NVIDIA A100 GPUs. We utilized the AdamW optimizer with a maximum learning rate of $1 \times 10^{-5}$, a cosine decay scheduler, and a 5\% warmup ratio. To handle long contexts, we enabled DeepSpeed ZeRO Stage 2 and set the per-device batch size to 1 with 16 gradient accumulation steps, resulting in an effective batch size of 128.

    \item \textbf{\redapo ~Fine-tuning}: Subsequently, we applied our proposed DAPO algorithm (implemented via the VeRL framework) on a single compute node equipped with 8 NVIDIA H200 GPUs. The model was optimized using AdamW with a learning rate of $1 \times 10^{-6}$ and a global batch size of 256, utilizing FSDP for model parallelism. To stabilize training and reduce variance, we adopted the ``Clip-Higher'' strategy with a DAPO clip range of $[0.2, 0.28]$, a KL penalty coefficient of $0.01$, and a rollout size of $N=8$. 
    
    Thanks to the SFT cold-start, format violations were rare; however, if the model generated unparseable or invalid JSON during RL exploration, we assigned a strict accuracy reward of 0, effectively teaching the model to maintain valid syntax autonomously. 
    
    The total RL training consisted of 870 steps (approximately 5 epochs on the full dataset). Due to the substantial computational cost of RL training on the VLM (requiring approximately 70 hours on a node with 8 $\times$ NVIDIA H200 GPUs), we were restricted to a single full-scale training run. Therefore, we did not perform multi-run statistical variance analysis.
\end{itemize}

\subsubsection{Baselines}
We evaluated the zero-shot performance of current mainstream general VLMs using identifiers as visual prompt, including GPT-4o~\cite{hurst2024gpt}, Gemini 3.0 Pro~\cite{gemini3pro2026}(with all metrics assessed in February 2026), Qwen2.5-VL-7B~\cite{bai2025qwen2}, and Intern-S1~\cite{bai2025intern}. Additionally, we tested specialized models for the RxnDP task, primarily RxnScribe~\cite{qian2023rxnscribe}, RxnIM~\cite{RxnIm}, and the BIVP version of RxnCaption-VL~\cite{song2025rxncaption}.

\subsection{Main Results}
\subsubsection{Strategies and Prior Knowledge}
The choice of visual prompt strategy is the primary determinant of model performance. As shown in Table 4, IdtVP consistently demonstrates a superior advantage over BIVP and BROS. Notably, IdtVP effectively activates the latent chemical reasoning and structural prior knowledge of large-scale foundation models like Gemini 3.0 Pro. For instance, Gemini 3.0 Pro (IdtVP) achieves a state-of-the-art (SOTA) Soft Match F1 of 86.5\% on the historical ScannedRxn dataset, even in a zero-shot setting. While BROS struggles with complex, real-world layouts and BIVP provides only coarse spatial cues, IdtVP offers fine-grained identifier-level guidance. Our RxnID+RL further leverages this strategy to reach a 64.4\% Hybrid Match F1 on RxnCaption-15k-test, surpassing all open-source baselines and competitive closed-source models such as GPT-4o~\cite{hurst2024gpt}.

\subsubsection{\redapo ~Performance and Stability}

While \redapo ~significantly improves in-distribution precision when applied to both BIVP and IdtVP strategies, the out-of-distribution stability of the resulting models varies sharply. To emphasize the impact of the underlying parsing strategies, we hereafter denote RxnCaption with \redapo\ as BIVP+RL, and RxnID with \redapo ~as IdtVP+RL. On the RxnCaption-15k-test set, both BIVP+RL and IdtVP+RL exhibit marked performance gains. However, when facing the severe domain shift of ScannedRxn, BIVP+RL often collapses into degenerate token repetition, leading to parsing failures and performance drops. Conversely, IdtVP+RL maintains high stability and avoids such looping hallucinations; the robust structural anchor provided by IdtVP facilitates reinforcement learning optimization, achieving a Hybrid Match F1 of 56.3\% on scanned data.

Besides, it is notably that RL nearly eliminates the performance gap between BIVP and IdtVP on in-distribution test sets. This directly proves our analysis in \cref{sec:re3_dapo} and \cref{tab:sft_inconsistency_breakdown} regarding \textbf{serialization noise}. While IdtVP naturally incurs fewer image-level and reaction-level serialization errors caused by token-level optimized SFT, reaction-level RL optimization effectively neutralizes this discrepancy, allowing BIVP to overcome its initial serialization penalty.

\subsubsection{Generalization and Scale}
The evaluation on RxnScribe-test reveals limited gains (e.g., 75.9\% Hybrid Match F1 for RxnCaption+RL), likely due to its small sample size (138 images) and inherent annotation inconsistencies, where minor score variations lack statistical significance. In contrast, the larger RxnCaption-15k-test set and the historical ScannedRxn dataset provide a more rigorous assessment of generalization. The latter highlights the severe limitations of synthetic-only training, as seen in RxnIM’s low 22.8\% Hybrid-F1, whereas our RL-refined models demonstrate the ability to capture universal reaction logic across divergent document eras.

\subsection{Ablation and Analysis}

\subsubsection{Influence of Identifier Detector}

\input{Tables/detector_comparison}
\cref{tab:detector_comparison} demonstrates that Mid-Mapper, as a lightweight model distilled from Gemini, achieves performance \textbf{comparable} to its teacher model. For instance, when evaluating with IdtVP+RL, the Soft Match F1 gap between Mid-Mapper and teacher model is merely 1.1\%, confirming the effectiveness of our distillation. Interestingly, using human-annotated ground truth as detector leads to slightly lower performance than using either Gemini or Mid-Mapper. Since IdtVP+RL was trained on Gemini-generated data, the natural inconsistencies in human labeling style and formatting essentially act as noise. Mid-Mapper avoids these by inheriting the teacher's stable output patterns, resulting in better compatibility with downstream tasks.
\input{Tables/stage_loss}
\subsubsection{Stage-wise Error Analysis}
To isolate performance bottlenecks, we perform a stage-wise error analysis on our top-performing RxnID+RL model (\cref{tab:stage_loss}). \texttt{MolYOLO Err.} and \texttt{Mid-Mapper Err.} represent the upper-bound loss from upstream detectors during visual prompt generation, assuming an ideal RxnDP VLM. \texttt{Assign Error} (Soft-F1) and \texttt{Textual Error} (Hybrid-F1) denote losses from incorrect index assignments and OCR failures, respectively.

Quantitative results show that \textbf{early-stage detectors contribute minimally} to overall error. However, on blurrier ScannedRxn, both MolYOLO and OCR become significant bottlenecks. Otherwise, RxnDP VLM remains the main bottleneck, struggling with complex graph topologies (e.g., dense cyclic flows) and unconventional visual symbols (e.g., lightbulb or lightning icons). Even with the cascading losses typical of multi-stage designs, \textbf{RxnID still surpass the trained, end-to-end BROS model} by roughly 10\% on all three datasets. Fine-grained metrics and error examples are provided in Appendix D.

\subsubsection{Influence of Reward Function}
\begin{figure}[h]
    \centering
    \includegraphics[width=\textwidth]{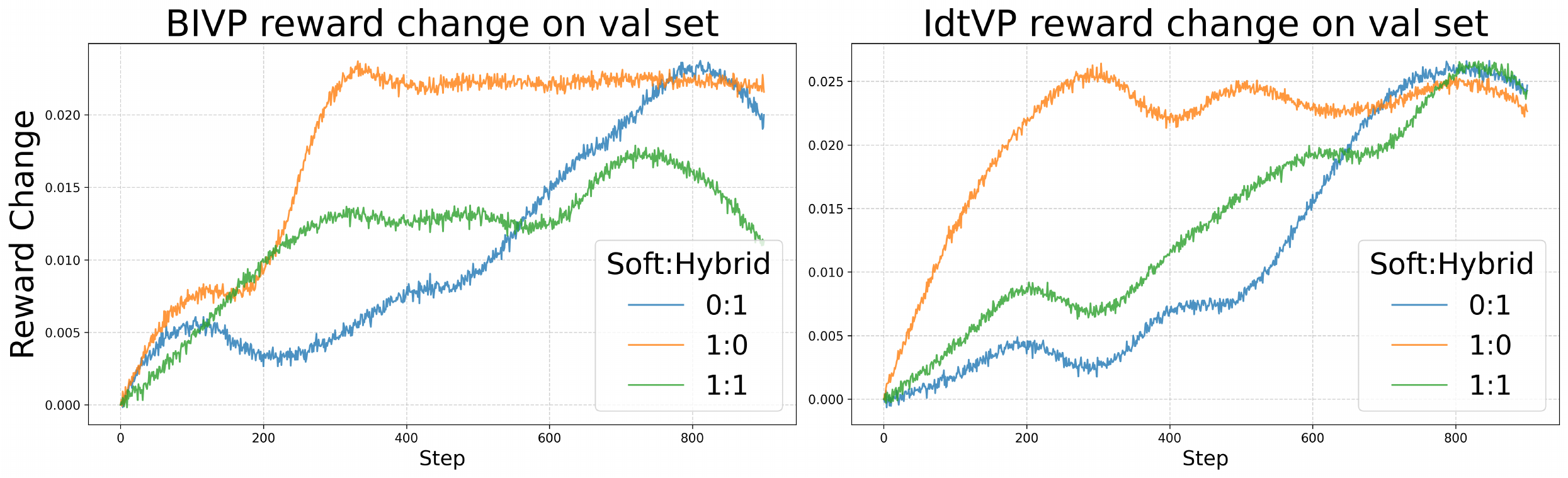}
    \caption{\textbf{Reward $\Delta$ change on the val dataset.} The Y-axis denotes the reward improvement relative to the initial step (Step 0).}
    \label{fig:reward_change}
\end{figure}
We ablated Soft:Hybrid reward ratios (1:0, 0:1, 1:1), monitoring reward $\Delta$ change on the val dataset and evaluating final performance on RxnCaption-15k-test.
\begin{itemize}
    \item \textbf{Training Dynamics}: As shown in~\cref{fig:reward_change}, the Soft-dominant (1:0) setting facilitates rapid initial convergence but plateaus early due to relaxed criteria. In contrast, the Hybrid-dominant (0:1) setting shows a slower start but maintains stronger upward momentum in later stages.
    \begin{table}[h]
    \centering
    \caption{\textbf{Quantitative results on RxnCaption-15k-test.} Comparison across reward ratios.}
    \label{tab:reward_function}
    \begin{tabular}{llcc}
        \toprule
        \multirow{2}{*}{\textbf{Ratio}} & \multirow{2}{*}{\textbf{Strategy}} & \multicolumn{2}{c}{\textbf{RxnCaption-15k-test}} \\ \cmidrule(lr){3-4}
         & & \textbf{Soft-F1} & \textbf{Hybrid-F1} \\ \midrule
        \multirow{2}{*}{0 : 1} & BIVP+RL & 72.5 & \textbf{64.5} \\
                               & IdtVP+RL  & 73.6 & \underline{64.4} \\ \midrule
        \multirow{2}{*}{1 : 0} & BIVP+RL & 73.8 & 62.7 \\
                               & IdtVP+RL  & \underline{74.2} & 62.8 \\ \midrule
        \multirow{2}{*}{1 : 1} & BIVP+RL & 73.6 & 64.1 \\
                               & IdtVP+RL  & \textbf{74.5} & \underline{64.4} \\ \bottomrule
    \end{tabular}
\end{table}
    \item \textbf{Final Performance}: As show in~\cref{tab:reward_function}, the balanced (1:1) configuration achieves the most stable trajectory and best generalization on RxnCaption-15k-test. It effectively combines the early-stage guidance of Soft Match with the precise optimization constraints of Hybrid Match, leading to the best overall performance across both models.
\end{itemize}

\subsection{IdtVP's Exclusive Cross-Modal Extensibility}
\label{sec:crossmodel}

\begin{figure}[h]
    \centering
    \includegraphics[width=\linewidth]{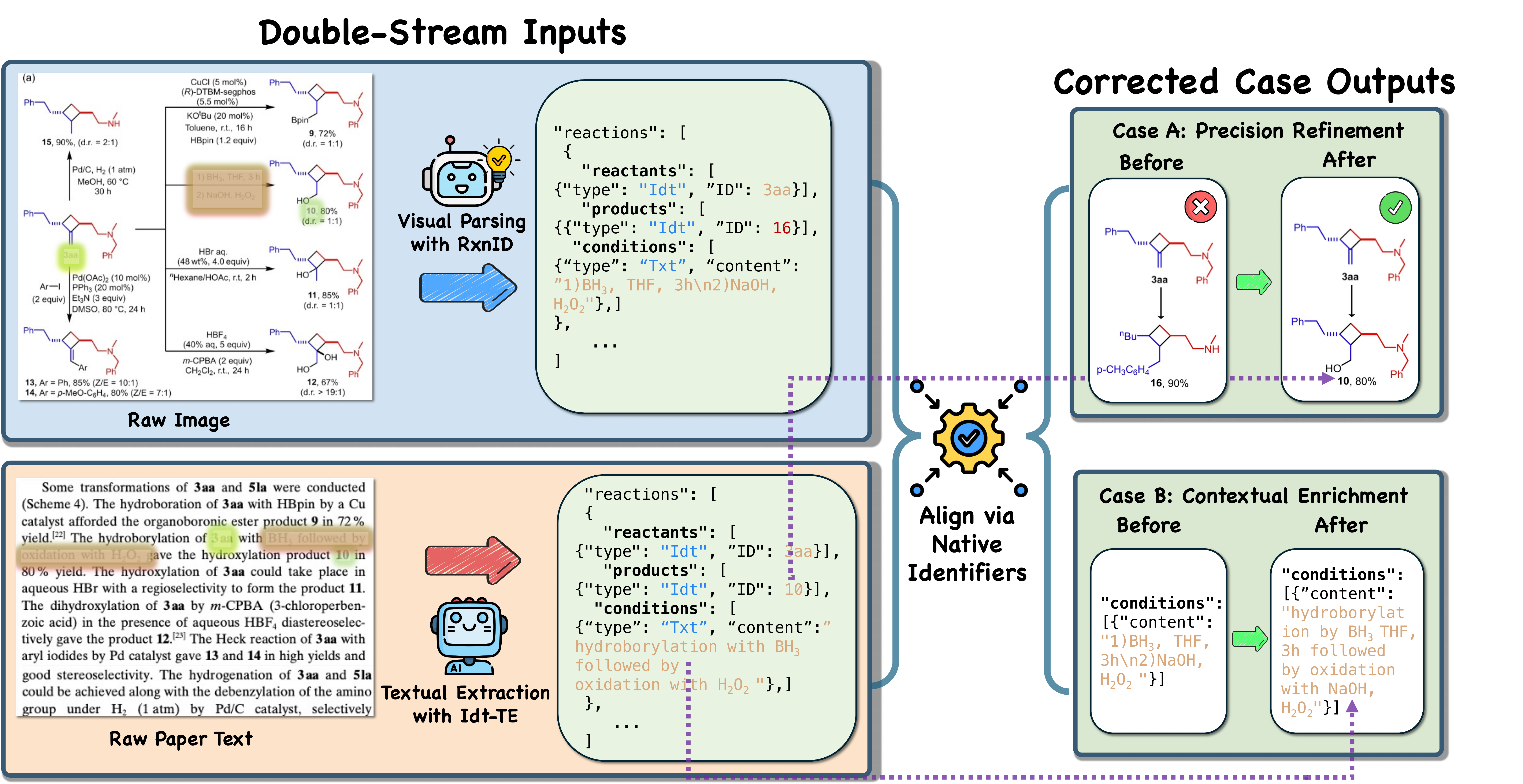}
    \caption{\textbf{Overview of the Cross-Modal Verification pipeline.} System processes double-stream inputs: visual parsing of raw diagrams via RxnID (top) and textual extraction from the manuscript via Idt-TE (bottom), enabling downstream applications such as \textbf{Precision Refinement} (Case A) and \textbf{Contextual Enrichment} (Case B).}
    \label{fig:Idt-TE}
\end{figure}

Unlike BIVP or Set-of-Mark which rely on isolated geometric anchors, IdtVP establishes a robust semantic bridge between vision and language. By adopting the \textbf{author's native identifiers} (e.g., \textbf{1a}) prevalent in both diagrams and text, IdtVP uniquely enables unified cross-modal indexing.

To demonstrate this exclusive advantage, we designed a \textbf{Cross-Modal Verification} pipeline (\cref{fig:Idt-TE}), which processes double-stream inputs via visual parsing (\textbf{RxnID}) and textual extraction (\textbf{Idt-TE}). By leveraging this shared vocabulary, the pipeline aligns LLM-extracted textual data with visual predictions seamlessly, without requiring any intermediate mapping. This exclusive capability enhances system robustness through:

\begin{itemize}
    \item \textbf{Precision Refinement (\cref{fig:Idt-TE}, Case A)}: Resolving visual ambiguities or OCR errors using textual context. This cross-referencing is inherently impossible for BIVP's arbitrary indices (e.g., Box \#1), which lack any semantic correlate in the manuscript text.
    \item \textbf{Contextual Enrichment (\cref{fig:Idt-TE}, Case B)}: Completing visual graphs with hidden attributes retrieved from the text. This positions IdtVP not merely as a parsing tool, but as the foundational bridge for a truly unified multimodal chemical knowledge extraction system.
\end{itemize}
Admittedly, while our Idt-TE pipeline comprehensively extracts rich reaction data from text, optimally applying these insights to the RxnDP task remains an open area for exploration, such as handling low-confidence refinements and determining if contextual enrichment exceeds the strict scope of pure diagram parsing. While exploring these integration strategies is left for future work, IdtVP's native identifiers uniquely provide the indispensable foundation for such unified extraction. Detailed Idt-TE's pipeline architectures, exact LLM prompts, each stage's output and representative qualitative case studies are provided in Appendix E.

%% file: Tables/exp.tex
\begin{table*}[t!]
\centering

\caption{\textbf{Model performance comparison on RxnScribe-test, RxnCaption-15k-test, and the new \textbf{ScannedRxn} datasets.} Best scores are in \textbf{bold}, second best are \underline{underlined}. \textbf{+RL} means post trained by \textbf{\redapo}, same as below.}
\label{tab:performance}

\begingroup 
\setlength{\aboverulesep}{0pt} 
\setlength{\belowrulesep}{0pt} 
\renewcommand{\arraystretch}{1.3} 
\setlength{\tabcolsep}{1.8pt}
\small 

\resizebox{\textwidth}{!}{
\begin{tabular}{@{}l l ccc ccc ccc ccc ccc ccc@{}}
\toprule
\multirow{3}{*}{\textbf{Model}} & \multirow{3}{*}{\textbf{Strategy}} & \multicolumn{6}{c}{\textbf{RxnScribe-test}} & \multicolumn{6}{c}{\textbf{RxnCaption-15k-test}} & \multicolumn{6}{c}{\textbf{ScannedRxn}} \\
\cmidrule(lr){3-8} \cmidrule(lr){9-14} \cmidrule(lr){15-20}
 & & \multicolumn{3}{c}{Hybrid Match} & \multicolumn{3}{c}{Soft Match} & \multicolumn{3}{c}{Hybrid Match} & \multicolumn{3}{c}{Soft Match} & \multicolumn{3}{c}{Hybrid Match} & \multicolumn{3}{c}{Soft Match} \\
\cmidrule(lr){3-5} \cmidrule(lr){6-8} \cmidrule(lr){9-11} \cmidrule(lr){12-14} \cmidrule(lr){15-17} \cmidrule(lr){18-20}
 & & \textbf{P} & \textbf{R} & \textbf{F1} & \textbf{P} & \textbf{R} & \textbf{F1} & \textbf{P} & \textbf{R} & \textbf{F1} & \textbf{P} & \textbf{R} & \textbf{F1} & \textbf{P} & \textbf{R} & \textbf{F1} & \textbf{P} & \textbf{R} & \textbf{F1} \\
\midrule
\multicolumn{20}{c}{\textit{Trained Models}} \\
\midrule
\rowcolor{gray!15} 
\textbf{RxnID+RL} & IdtVP & \textbf{76.3} & \underline{73.7} & \underline{75.0} & 87.2 & 84.6 & 85.9 & \textbf{65.5} & \textbf{63.4} & \textbf{64.4} & \underline{75.8} & 73.2 & \underline{74.5} & 60.2 & \underline{52.8} & \underline{56.3} & \underline{82.5} & \underline{70.8} & \underline{76.2} \\
\textbf{RxnID} & IdtVP & 75.8 & 73.5 & 74.6 & 87.2 & 84.5 & 85.6 & 64.2 & 61.1 & 61.2 & 75.2 & 70.6 & 72.8 & 58.1 & 51.4 & 54.5 & 80.5 & 69.1 & 74.4 \\
\rowcolor{gray!15} 
\textbf{RxnCaption+RL} & BIVP & \underline{76.0} & \textbf{75.8} & \textbf{75.9} & 86.8 & 87.1 & \underline{86.9} & \textbf{65.5} & \underline{62.9} & \underline{64.1} & 75.3 & 72.0 & 73.6 & 54.2 & 38.6 & 45.1 & 72.2 & 50.2 & 59.2 \\
\multirow{2}{*}{RxnCaption} & BIVP & 71.6 & 72.7 & 72.2 & 85.3 & 87.1 & 86.2 & 60.3 & 59.3 & 59.8 & 71.3 & 69.4 & 70.4 & 54.6 & 47.9 & 51.0 & 73.1 & 66.2 & 69.5 \\
 & BROS & 69.6 & 68.9 & 69.2 & 76.2 & 76.2 & 76.2 & 57.0 & 57.5 & 57.2 & 66.4 & 67.4 & 66.9 & 56.3 & 52.3 & 54.2 & 68.7 & 67.6 & 68.2 \\
\addlinespace[2pt]
\rowcolor{gray!15} 
RxnScribe\_w/15k & BROS & 72.4 & 69.1 & 70.7 & 84.1 & 81.7 & 82.8 & 61.2 & 38.7 & 47.4 & 72.1 & 44.7 & 55.2 & \underline{60.7} & 43.3 & 50.5 & 76.1 & 58.2 & 65.9 \\
RxnScribe & BROS & 72.3 & 66.2 & 69.1 & 83.8 & 76.5 & 80.0 & 47.4 & 27.6 & 34.9 & 62.1 & 36.4 & 45.9 & 57.6 & 42.8 & 49.1 & 75.5 & 59.7 & 66.7 \\
\rowcolor{gray!15} 
RxnIM & BROS & 71.0 & 70.1 & 70.5 & 79.2 & 74.7 & 76.9 & 48.8 & 30.3 & 37.4 & 52.9 & 32.8 & 40.5 & 43.1 & 15.5 & 22.8 & 48.6 & 23.4 & 31.6 \\
\midrule
\multicolumn{20}{c}{\textit{Open-source Models}} \\
\midrule
\rowcolor{gray!15} 
 & BIVP & 49.9 & 53.1 & 51.4 & 66.4 & 71.2 & 68.8 & 27.8 & 24.4 & 26.0 & 44.7 & 40.2 & 42.3 & 11.2 & 9.3 & 10.2 & 23.7 & 19.9 & 21.7 \\
\rowcolor{gray!15} 
\multirow{-2}{*}{Intern-S1} & IdtVP & 53.8 & 38.3 & 44.7 & 72.7 & 52.3 & 60.8 & 33.8 & 17.1 & 22.7 & 53.3 & 27.7 & 36.5 & 44.5 & 35.5 & 39.5 & 71.4 & 58.4 & 64.3 \\
\addlinespace[2pt]
\multirow{2}{*}{Qwen2.5-VL-7B} 
 & BIVP & 6.0 & 4.1 & 4.9 & 55.8 & 36.0 & 43.8 & 2.9 & 0.9 & 1.4 & 33.0 & 10.3 & 15.6 & 37.1 & 31.7 & 34.2 & 51.3 & 45.3 & 48.1 \\
 & IdtVP & 21.2 & 13.8 & 16.7 & 49.0 & 32.4 & 39.0 & 10.1 & 3.2 & 4.9 & 38.6 & 12.7 & 19.2 & 40.9 & 30.1 & 34.7 & 63.5 & 50.6 & 56.3 \\
\midrule
\multicolumn{20}{c}{\textit{Closed-source Models}} \\
\midrule
\rowcolor{gray!15} 
 & BIVP & 41.7 & 49.5 & 45.3 & 74.4 & \textbf{89.6} & 81.3 & 46.5 & 48.4 & 47.4 & 71.6 & \underline{75.3} & 73.4 & 18.3 & 17.8 & 18.6 & 51.2 & 49.5 & 50.3 \\
\rowcolor{gray!15} 
\multirow{-2}{*}{Gemini 3.0 Pro} & IdtVP & 72.0 & 71.4 & 71.7 & \textbf{87.9} & \underline{88.3} & \textbf{88.1} & 59.0 & 57.6 & 58.3 & \textbf{78.5} & \textbf{78.0} & \textbf{78.2} & \textbf{60.7} & \textbf{58.0} & \textbf{59.3} & \textbf{89.4} & \textbf{83.9} & \textbf{86.5} \\
\addlinespace[2pt]
\multirow{2}{*}{GPT-4o} 
 & BIVP & 19.7 & 25.3 & 22.1 & 47.9 & 59.1 & 52.9 & 14.0 & 16.0 & 15.0 & 27.5 & 29.9 & 28.7 & 10.9 & 8.9 & 9.8 & 27.9 & 19.3 & 22.8 \\
 & IdtVP & 37.7 & 37.8 & 37.7 & 66.3 & 67.4 & 66.8 & 20.9 & 17.3 & 18.3 & 48.8 & 36.7 & 41.9 & 41.8 & 30.6 & 35.3 & 65.8 & 52.2 & 58.2 \\
\addlinespace[2pt]
\rowcolor{gray!15} 
 & BIVP & 43.1 & 44.4 & 43.7 & 70.1 & 73.3 & 71.7 & 34.7 & 26.2 & 29.8 & 63.6 & 48.8 & 55.2 & 21.3 & 16.3 & 18.5 & 34.6 & 27.8 & 30.8 \\
\rowcolor{gray!15} 
\multirow{-2}{*}{Qwen-VL-Max} & IdtVP & 48.7 & 42.9 & 45.6 & 72.4 & 64.5 & 68.2 & 26.8 & 25.3 & 26.0 & 44.9 & 43.1 & 44.0 & 49.7 & 36.2 & 41.9 & 73.9 & 59.4 & 65.9 \\
\bottomrule
\end{tabular}
}
\endgroup
\vspace{-1em}
\end{table*}

%% file: Tables/detector_comparison.tex
\begin{table}[ht!]
    \centering
    \caption{\textbf{Performance comparison of different identifier detectors on the RxnCaption-15k-test set.} We evaluate the impact of detector precision on the Soft-F1 and Hybrid-F1 scores across different VLM prompting strategies.}
    \label{tab:detector_comparison}
    \begin{tabular}{llcc}
        \toprule
        \multirow{2}{*}{\textbf{Detector}} & \multirow{2}{*}{\textbf{VLM}} & \multicolumn{2}{c}{\textbf{RxnCaption-15k-test}} \\ \cmidrule(lr){3-4}
         & & \textbf{Soft-F1} & \textbf{Hybrid-F1} \\ \midrule
        \multirow{2}{*}{Mid-Mapper} & IdtVP+RL & 74.5 & \underline{64.4} \\
                                    & Gemini   & \underline{78.2} & 58.3 \\ \midrule
        \multirow{2}{*}{Gemini}     & IdtVP+RL & 75.6 & \textbf{64.7} \\
                                    & Gemini   & \textbf{78.6} & 60.2 \\ \midrule
        \multirow{2}{*}{Ground Truth}& IdtVP+RL & 73.6 & 63.1 \\
                                    & Gemini   & 77.9 & 58.0 \\ \bottomrule
    \end{tabular}
\end{table}

%% file: Tables/stage_loss.tex
\begin{table}[h]
    \centering
    \caption{\textbf{Main results on test set} (RS=RxnScribe, RC=RxnCaption-15k-test, SR=ScannedRxn). Met. and Drp.($\downarrow$) denote the Metric value and the performance drop relative to the upper bound, respectively.}
    \label{tab:stage_loss}
    
    \begingroup
    \setlength{\aboverulesep}{0pt}
    \setlength{\belowrulesep}{0pt}
    \renewcommand{\arraystretch}{1.3} 
    \begin{tabular}{lcccccc}
        \toprule
        \multirow{2}{*}{\textbf{Error Source}} & \multicolumn{2}{c}{\textbf{RS}} & \multicolumn{2}{c}{\textbf{RC}} & \multicolumn{2}{c}{\textbf{SR}} \\
        \cmidrule(lr){2-3} \cmidrule(lr){4-5} \cmidrule(lr){6-7}
        & Met. & Drp$\downarrow$ & Met. & Drp$\downarrow$ & Met. & Drp$\downarrow$ \\
        \midrule
        
        \rowcolor{gray!15} 
        \textit{Upper Bound} & 100.0 & - & 100.0 & - & 100.0 & - \\
        
        \textbf{MolYOLO Err.} & & & & & & \\
        (Ideal RxnDP VLM) & \multirow{-2}{*}{99.7} & \multirow{-2}{*}{\textbf{0.3}} & \multirow{-2}{*}{95.3} & \multirow{-2}{*}{\textbf{4.7}} & \multirow{-2}{*}{87.1} & \multirow{-2}{*}{\textbf{12.9}} \\
        
        \rowcolor{gray!15} 
        \textbf{Mid-Mapper Err.} & & & & & & \\
        \rowcolor{gray!15} 
        (Ideal RxnDP VLM) & \multirow{-2}{*}{98.5} & \multirow{-2}{*}{\textbf{1.2}} & \multirow{-2}{*}{93.8} & \multirow{-2}{*}{\textbf{1.5}} & \multirow{-2}{*}{84.7} & \multirow{-2}{*}{\textbf{2.4}} \\
        
        \textbf{Assign Error} & & & & & & \\
        (Soft-F1) & \multirow{-2}{*}{85.9} & \multirow{-2}{*}{\textbf{12.6}} & \multirow{-2}{*}{74.5} & \multirow{-2}{*}{\textbf{19.3}} & \multirow{-2}{*}{76.2} & \multirow{-2}{*}{\textbf{8.5}} \\
        
        \rowcolor{gray!15} 
        \textbf{Textual Error} & & & & & & \\
        \rowcolor{gray!15} 
        (Hybrid-F1) & \multirow{-2}{*}{75.0} & \multirow{-2}{*}{\textbf{10.9}} & \multirow{-2}{*}{64.4} & \multirow{-2}{*}{\textbf{10.1}} & \multirow{-2}{*}{56.3} & \multirow{-2}{*}{\textbf{19.9}} \\
        
        \bottomrule
    \end{tabular}
    \endgroup
\end{table}

%% file: sec/6_conclusion.tex
\section{Conclusion}
In this paper, we propose RxnID, a framework leveraging identifiers as visual prompts to activate VLM domain priors for precise chemical reaction parsing. To overcome optimization misalignment and the forced serialization noise inherent in SFT, we introduce $Re^3$-DAPO, a reinforcement learning strategy that utilizes permutation-invariant rewards to enforce sequence-level correctness. Beyond visual parsing, our IdtVP strategy establishes a unique semantic bridge that enables a Cross-Modal Verification pipeline for fine-grained precision refinement and contextual enrichment. We will further explore this cross-modal direction in future work. Finally, we release ScannedRxn to evaluate model robustness on historical literature. Extensive experiments confirm that RxnID and \redapo~ strategy significantly outperforms existing state-of-the-art methods, offering a robust and extensible solution for digitizing chemical archives.

%% file: sec/sup.tex
\input{sec/related_work}
\input{sec/I-Det}
\input{sec/Scannedrxn}
\input{sec/RL}

\input{sec/Idt-TE-details}
\clearpage
\newpage

%% file: sec/related_work.tex
\section{Related Work}
\label{sec:related-work}

\subsection{VLMs for Document Parsing and Visual Prompting}

Document understanding~\cite{wei2024vary,liu2024textmonkey,liu2024textmonkey,li2025uni} has largely transitioned from OCR-centric pipelines to vision-language paradigms. Early encoder-decoder models like Donut~\cite{kim2022ocr} and Nougat~\cite{blecher2023nougat} extracted structured text directly from images. General-purpose VLMs, including LLaVA~\cite{liu2023visual} and Qwen-VL~\cite{bai2023qwen, bai2023qwenvlversatilevisionlanguagemodel, wang2024qwen2}, advanced multimodal comprehension via instruction tuning. Subsequent efforts such as Vary~\cite{wei2024vary}, and TextMonkey~\cite{liu2024textmonkey} optimized high-resolution inputs to capture fine-grained details in dense text. Despite these improvements, spatial localization remains a bottleneck for scientific diagrams. When processing intertwined symbols and structures, general VLMs frequently suffer from spatial misalignment and structural hallucinations~\cite{bai2024hallucination, wu2026firered}. Visual prompting techniques like Set-of-Mark (SoM)~\cite{yang2023set} and ViP-LLaVA~\cite{cai2024vip} address this by superimposing explicit visual cues (e.g., bounding boxes or points) to trigger zero-shot grounding. However, these methods are primarily designed for salient objects in natural images. They struggle to resolve the highly dense, fine-grained associations between chemical structures and alphanumeric identifiers common in scientific literature.

\subsection{Chemical Reaction Extraction and Parsing}

The extraction of chemical reactions has evolved from text mining to visual diagram parsing. Text-based extraction progressed from rule-based systems~\cite{schneider2016big, schneider2016s} to Transformer architectures~\cite{guo2021automated, zhong2023reaction} and LLMs~\cite{vangala2024suitability}. Correspondingly, visual parsing shifted from detection-based pipelines like ReactionDataExtractor 2.0~\cite{wilary2023reactiondataextractor} to generative sequence modeling. RxnScribe~\cite{qian2023rxnscribe} established a baseline using Pix2Seq, and RxnIM~\cite{RxnIm} scaled this approach with synthetic data. Recently, RxnCaption~\cite{song2025rxncaption} introduced visual cues inspired by general prompting methods~\cite{yang2023set, zhou2024image}. Nevertheless, these approaches treat chemical diagrams as generic visual signals, ignoring the domain-specific topological constraints of chemical structures. Consequently, end-to-end multimodal models are prone to failures in complex coreference resolution between molecules and their identifiers~\cite{guoend}. When encountering missing labels, complex hyphenations (e.g., (E)-3), or dense visual occlusions, the lack of explicit logical validation in the parsing process frequently leads to cascade errors.

\subsection{Reinforcement Learning for Multimodal Reasoning}

Reinforcement learning (RL) is increasingly utilized to elicit complex reasoning in large models. Following foundation models like DeepSeek-R1~\cite{guo2025deepseek} and GPT-o1~\cite{jaech2024openai}, studies such as Tulu 3~\cite{lambert2024tulu} demonstrated that post-training RL effectively enforces internal logical consistency~\cite{chu2025sft, luo3deepscaler}. This paradigm is rapidly expanding into the multimodal domain. Recent works including Vision-R1~\cite{huang2025vision}, Perception-R1~\cite{yu2025perception}, and VLM-R1~\cite{shen2025vlm} have extended test-time computation and "slow-thinking" capabilities to visual perception, improving spatial localization and long-chain reasoning~\cite{wen2025reinforcement}. Concurrently, frameworks like Open-VLThinker~\cite{deng2025openvlthinker}, Dr. GRPO~\cite{liu2025understanding}, and DAPO~\cite{yu2025dapo} have explored the stability of multimodal alignment.

Mainstream multimodal RL heavily relies on costly human feedback (RLHF) or AI-generated preferences (RLAIF)~\cite{yu2024rlhf, zhang2502mm}. In contrast, scientific diagrams possess objective, strict domain rules, such as mass conservation, valence constraints, and topological cycle consistency~\cite{li2026agentic, levine2026tssr}. Recent investigations into verifiable rewards~\cite{wang2025rlver} indicate that deterministic, rule-based feedback provides a more efficient and stable signal for model convergence, directly mitigating hallucinations. Despite this, integrating verifiable objective constraints with visual prompting for the rigorous topological parsing of scientific diagrams has not been systematically investigated. Our work formulates these deterministic chemical rules into a verifiable reward mechanism, providing a reliable training signal for visual prompt-guided parsing without relying on human preference annotations.

%% file: sec/I-Det.tex
\section{Details of Mid-Mapper}
\label{midmapper}
\subsection{Mid-Mapper's Data Curation Details}
Prompt for Data Distillation To curate high-quality training data for the Mid-Mapper model, we employed Gemini 3 Pro for knowledge distillation on visually augmented images. A structured system prompt was designed to guide the model in mapping ``sequential box indice'' to ``semantic chemical identifiers'' and inferring consistent virtual identifiers for unlabeled molecules. The full prompt is provided in \cref{tab:prompt_extraction}.
\input{Tables/I-Det_prompt}

\subsection{Adaptive Rendering of Virtual Identifiers}

To explicitly visualize virtual identifiers while maintaining visual coherence, we developed an Ink-Aware Adaptive Rendering Algorithm. The algorithm first employs OCR to analyze existing labels, dynamically inferring the naming schema and optimal font size to match the original style. Subsequently, it utilizes a priority-based placement strategy that searches for whitespace around molecular bounding boxes, enforcing a pixel-level ink density check to strictly prevent occlusion of chemical structures. For dense layouts, a robust fallback mechanism is triggered, employing iterative font scaling and spiral pixel scanning to locate viable placement coordinates without compromising the semantic integrity of the original diagram.

%% file: Tables/I-Det_prompt.tex
\begin{table*}[t]
\centering
\caption{Prompt used for identifier extraction.}
\label{tab:prompt_extraction} 
\begin{promptbox}{Prompt used for identifier extraction}
    \small \sffamily
    You are an AI assistant specialized in extracting the identifiers of molecular formulas. Follow the instructions below precisely:

    \begin{enumerate}[label=\arabic*., leftmargin=*, nosep, font=\bfseries, itemsep=3pt, topsep=3pt]
        \item \textbf{Identifier Definition:}
        \begin{itemize}[label=\tiny$\bullet$, nosep]
            \item An identifier is an explicit label used to refer to a molecular formula in the diagram.
            \item It is typically shown next to the molecular formula, in boldface, and usually consists of short numbers or letters (e.g., "1", "2", "a", "b", "2c").
            \item Any text surrounding a molecule but not in bold \textbf{must not} be an identifier.
        \end{itemize}

        \item \textbf{Identifier Processing Rules:}
        \begin{itemize}[label=\tiny$\bullet$, nosep]
            \item Each molecule is enclosed in a blue bounding box, with a box index (black background, white text) in the top-left corner.
            \item When bold text is connected by hyphens "-", such as "(E)-3, (+)-3f", please extract them completely as identifier.
        \end{itemize}

\item \textbf{Missing Identifier Handling:}
        \begin{itemize}[label=\tiny$\bullet$, nosep]
            \item If no clearly bolded identifier is nearby, assign a new identifier following the existing style.
            \item Set a special field \texttt{"is\_virtual"} to \texttt{true} in this item.
        \end{itemize}
    \end{enumerate}

    \vspace{2mm}
    \hrule
    \vspace{2mm}
    
    \noindent\textbf{Example Output:}
    \begin{lstlisting}[basicstyle=\ttfamily\scriptsize, columns=fullflexible, breaklines=true]
[
  { "mol_index": "1", "identifier": ["1"] },
  { "mol_index": "2", "identifier": ["2a", "2c"] },
  { "mol_index": "3", "identifier": ["3c"], "is_virtual": true }
]
    \end{lstlisting}
\end{promptbox}
\end{table*}

%% file: sec/Scannedrxn.tex
\section{Details of ScannedRxn}
\label{sec:details_of_SR}

\subsection{Details of Construction Pipeline}
\label{sec:details_of_SR_construction}

To systematically evaluate the model's generalization capabilities on historical chemical archives, we constructed the \textbf{ScannedRxn} dataset. Unlike modern born-digital literature, historical publications spanning from the 1950s to the 1990s present unique visual challenges, including severe scanning noise, legacy layout conventions, and typewriter-style fonts. The dataset consists of 200 carefully curated reaction diagrams, evenly distributed across four topological complexities: single, multi-line, tree, and cyclic reactions (50 samples each).

\begin{figure}[htbp]
  \centering
  \includegraphics[width=\textwidth]{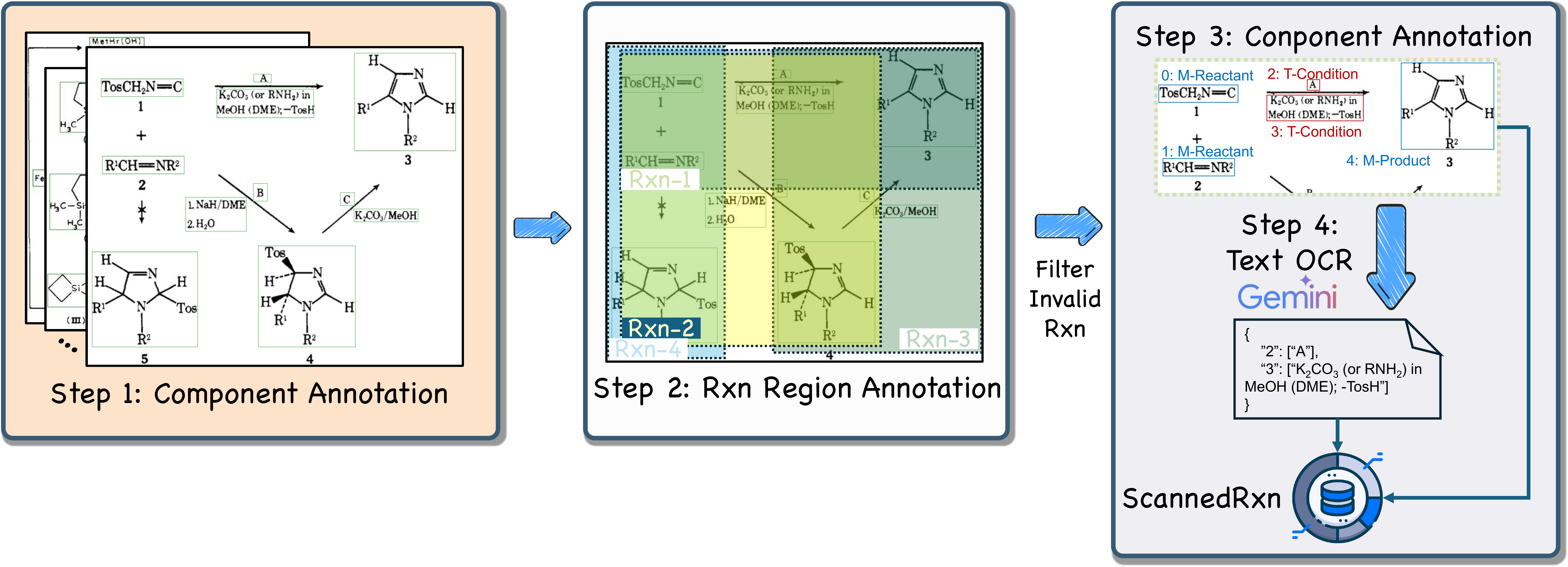} 
  \caption{Overview of the annotation pipeline for the ScannedRxn dataset.}
  \label{fig:scannedrxn_pipeline}
\end{figure}

To ensure consistency with modern benchmarks while accommodating the artifacts of historical scans, the annotation pipeline for ScannedRxn adapts the foundational RxnCaption~\cite{song2025rxncaption} methodology into four rigorous steps, as illustrated in \cref{fig:scannedrxn_pipeline}:

\begin{itemize}
    \item \textbf{Step 1: Component Annotation.} We first localize all fundamental elements within the scanned diagrams. Bounding boxes are meticulously applied to isolate individual molecular structures, reaction condition text blocks.
    
    \item \textbf{Step 2: Reaction Region Annotation.} We use bounding regions (polygons) to group the localized components into distinct, independent chemical reactions (e.g., Rxn-1, Rxn-2). This step delineates the boundaries of overlapping or densely packed reaction pathways.
    
    \item \textbf{Step 3: Component Role Annotation.} Prior to this step, it is necessary to filter out invalid reaction regions. This includes non-generative pathways (such as Rxn-4 in Step 2), energy conversion diagrams, and other non-conventional organic reactions. Within each valid defined reaction region, we then assign a specific chemical role to every localized component. Here, \texttt{M-xx} denotes a Molecular Structure and \texttt{T-xx} represents Text Content.
    
    \item \textbf{Step 4: Text OCR.} For this step, we leverage the advanced vision-language capabilities of Gemini 3 Pro to perform automated Optical Character Recognition (OCR). The model reads the localized condition blocks and outputs the recognized text into a structured JSON format.
\end{itemize}

\subsection{Examples of ScannedRxn}
\label{sec:exp_of_SR}
\begin{figure}[htbp]
  \centering
  \includegraphics[width=\textwidth]{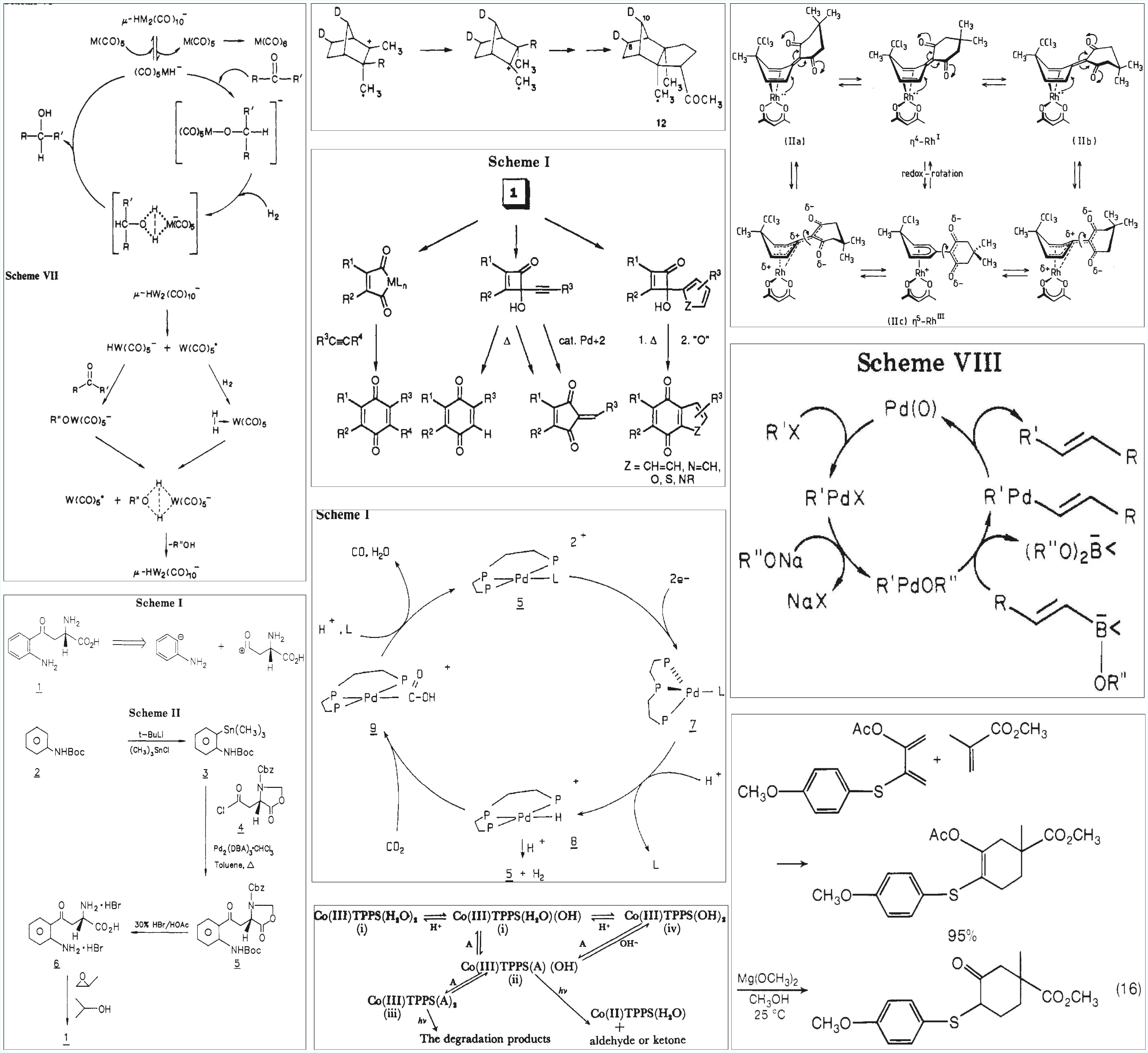} 
  \caption{Visualization of sample data from the ScannedRxn dataset.}
  \label{fig:scannedrxn_examples}
\end{figure}

\cref{fig:scannedrxn_examples} illustrates representative reaction diagram samples from the ScannedRxn dataset. Extracted from historical chemical literature published between the 1950s and 1990s, these images encompass four reaction topologies: single, multi-line, tree, and cyclic. Unlike the clean digital vector graphics found in modern publications, these samples exhibit the visual characteristics of early printed media, including heavy scanning artifacts, low contrast, typewriter fonts, and non-standard legacy layouts. As shown, high-density noise and irregular character spacing frequently obscure the visual boundaries between molecular structures and their corresponding identifiers. These historical features pose distinct challenges to existing end-to-end parsing models, validating the utility of the dataset for evaluating cross-era model generalization.

%% file: sec/RL.tex
\section{Detailed Error Analysis and Case Studies}
\label{sec:detailed_exp}
\subsection{Comprehensive Evaluation Metrics}
\label{sec:metrics}
Since the BROS baseline does not include a Reinforcement Learning (RL) stage, we remove the influence of RL and focus exclusively on the Supervised Fine-Tuning (SFT) phase in this section to explicitly demonstrate stage errors. Accordingly, we present the fine-grained F1 scores of three strategies (BROS, BIVP, and IdtVP) using SFT on Qwen2.5-VL~\cite{bai2025qwen2}, alongside the zero-shot performance of Gemini 3.0 Pro~\cite{gemini3pro2026}. 

\begin{figure*}[htbp]
  \centering
  \includegraphics[width=\textwidth]{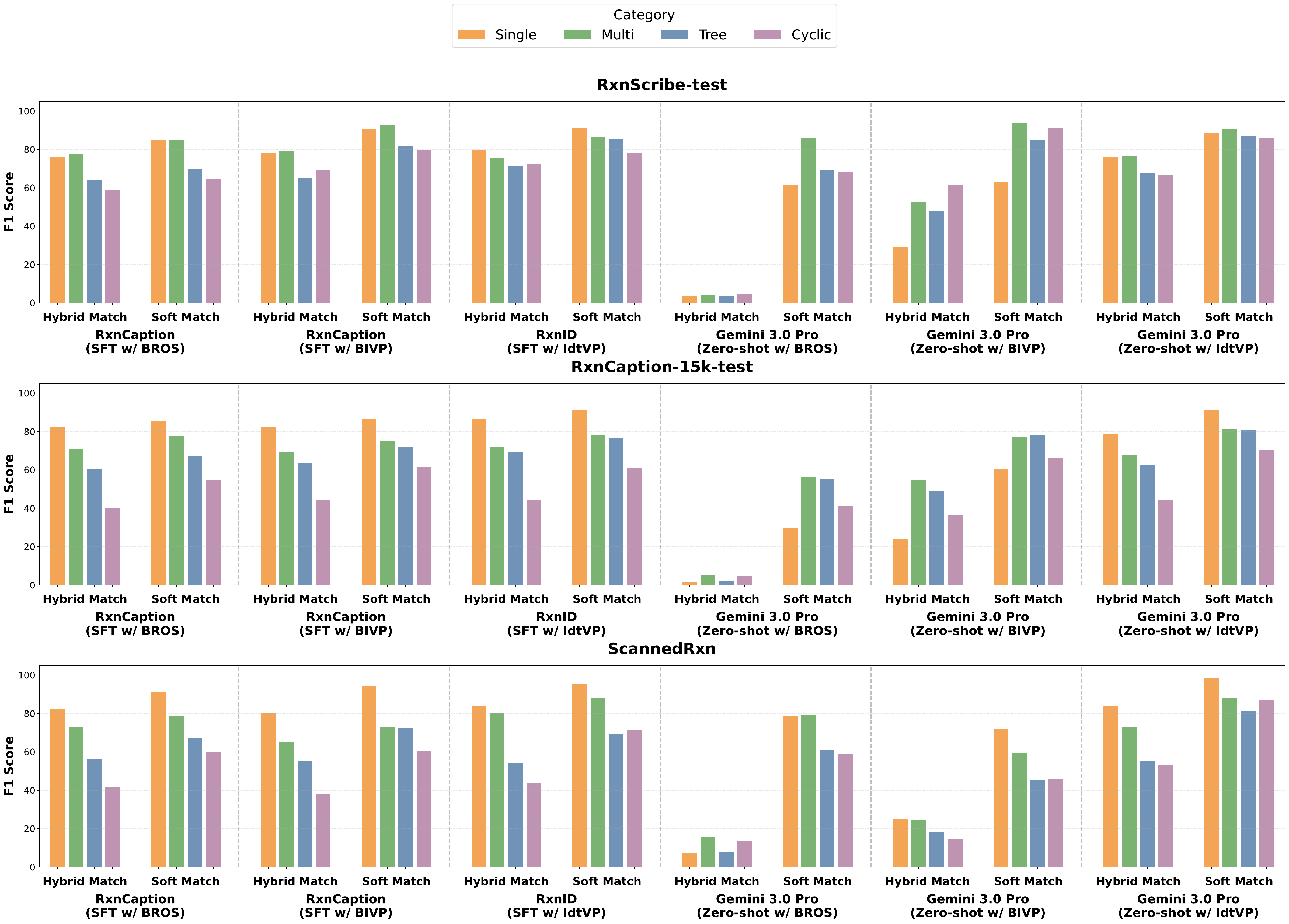} 
  \caption{\textbf{Fine-grained performance comparison across different reaction diagram types (Single, Multi-line, Tree, and Cyclic).} We evaluate the F1 scores (Hybrid and Soft Match) by performing Supervised Fine-Tuning (SFT) on Qwen2.5-VL and testing the zero-shot capabilities on Gemini 3.0 Pro. Both are evaluated under three representation strategies: BROS, BIVP, and IdtVP.}
  \label{fig:fine_grained_comparison}
\end{figure*}

As illustrated in~\cref{fig:fine_grained_comparison}, IdtVP demonstrates a clear advantage across most datasets and diagram types. Under the SFT setting, while the BROS baseline suffers severe performance degradation on non-linear structures (e.g., \textit{Tree} and \textit{Cyclic}), IdtVP maintains robust performance. Although IdtVP slightly underperforms BIVP on certain categories within the RxnScribe-test set, this discrepancy is primarily attributed to the extreme scarcity of such complex samples in this dataset. Overall, IdtVP exhibits superior generalization in parsing diverse reaction layouts.

Moreover, IdtVP effectively elicits the pre-trained reasoning capabilities of Gemini 3.0 Pro. While the text-based BROS strategy shows limited capability in interpreting complex non-linear topologies, IdtVP provides explicit visual anchoring for reaction entities. Benefiting from this LLM-friendly representation, Gemini 3.0 Pro achieves excellent zero-shot performance across all four diagram types. It not only addresses the spatial understanding limitations of BROS but also demonstrates a substantial performance margin over BIVP.

\subsection{Qualitative Error Gallery}
\label{sec:error_gallery}
\begin{figure*}[h]
  \centering
  \includegraphics[width=\textwidth]{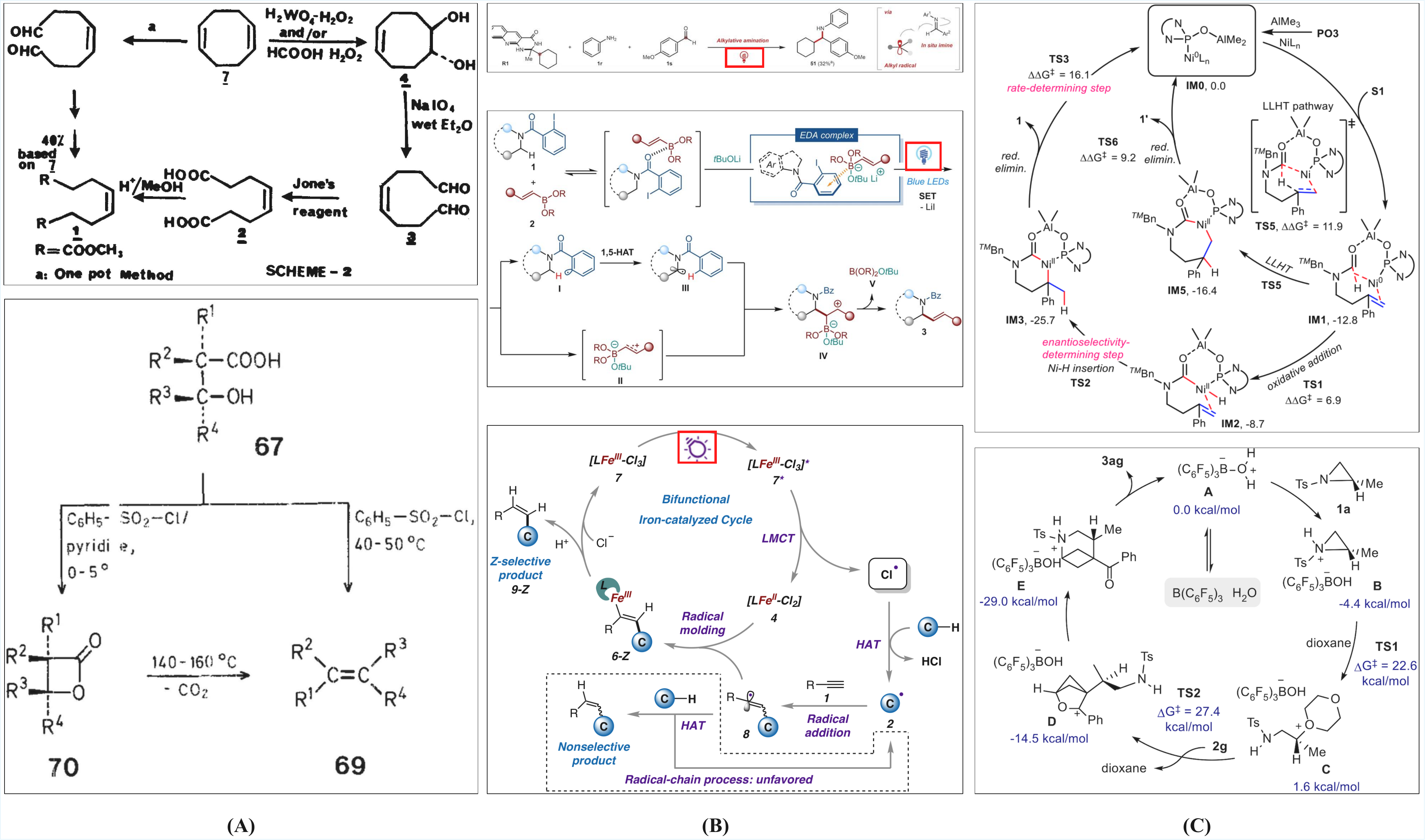} 
  \caption{\textbf{Typical error cases of RxnID.} We categorize the common failure modes into three types: (A) Low-resolution diagrams, where degraded visual quality frequently leads to inaccurate OCR predictions. (B) The presence of unconventional graphical icons (e.g., lightbulbs or sun symbols, highlighted by red boxes). Since these are neither standard chemical structures nor text, the model struggles to interpret them correctly. (C) Highly complex \textit{Cyclic} diagrams containing a dense concentration of reaction steps, which challenge the model's spatial reasoning capabilities.}
  \label{fig:error_cases}
\end{figure*}
To better understand the limitations of RxnID, we qualitatively analyze its typical error cases shown in~\cref{fig:error_cases}. First, as illustrated in (A), many scanned or printed chemical diagrams suffer from low resolution. In such degraded images, characters often blur or stick together; since Vision-Language Models (VLMs) are highly sensitive to image resolution, this directly leads to incorrect OCR predictions. Second, (B) highlights failures caused by out-of-scope graphical icons, such as a lightbulb or a sun (marked by red boxes). Because the current parsing scope of RxnDP strictly covers chemical formulas and standard text, these unconventional visual symbols fall outside the model's recognition capabilities. Addressing this would require more fine-grained annotations for visual elements in future datasets. Finally, (C) demonstrates the challenges posed by highly complex \textit{Cyclic} diagrams. These structures contain a massive amount of intertwined reaction information. Unlike basic linear reactions that follow straightforward reading orders (e.g., left-to-right or top-to-bottom), cyclic diagrams feature cluttered arrows and non-standard directionality, which easily confuse the VLM and lead to spatial parsing errors.

%% file: sec/Idt-TE-details.tex
\section{Details of Idt-TE}
\label{sec:details_of_Idt-TE}

As discussed in the main text, the inherent advantage of our IdtVP framework is the utilization of the author's native identifiers, which establish a robust semantic bridge between visual diagrams and manuscript text. To fully exploit this shared vocabulary, we introduce the \textbf{(Identifier-based Textual Extraction) Idt-TE} pipeline. The extraction process is decomposed into three rigorous phases to ensure high-fidelity alignment with visual predictions.

\begin{figure}[h]
    \centering
    \includegraphics[width=\textwidth]{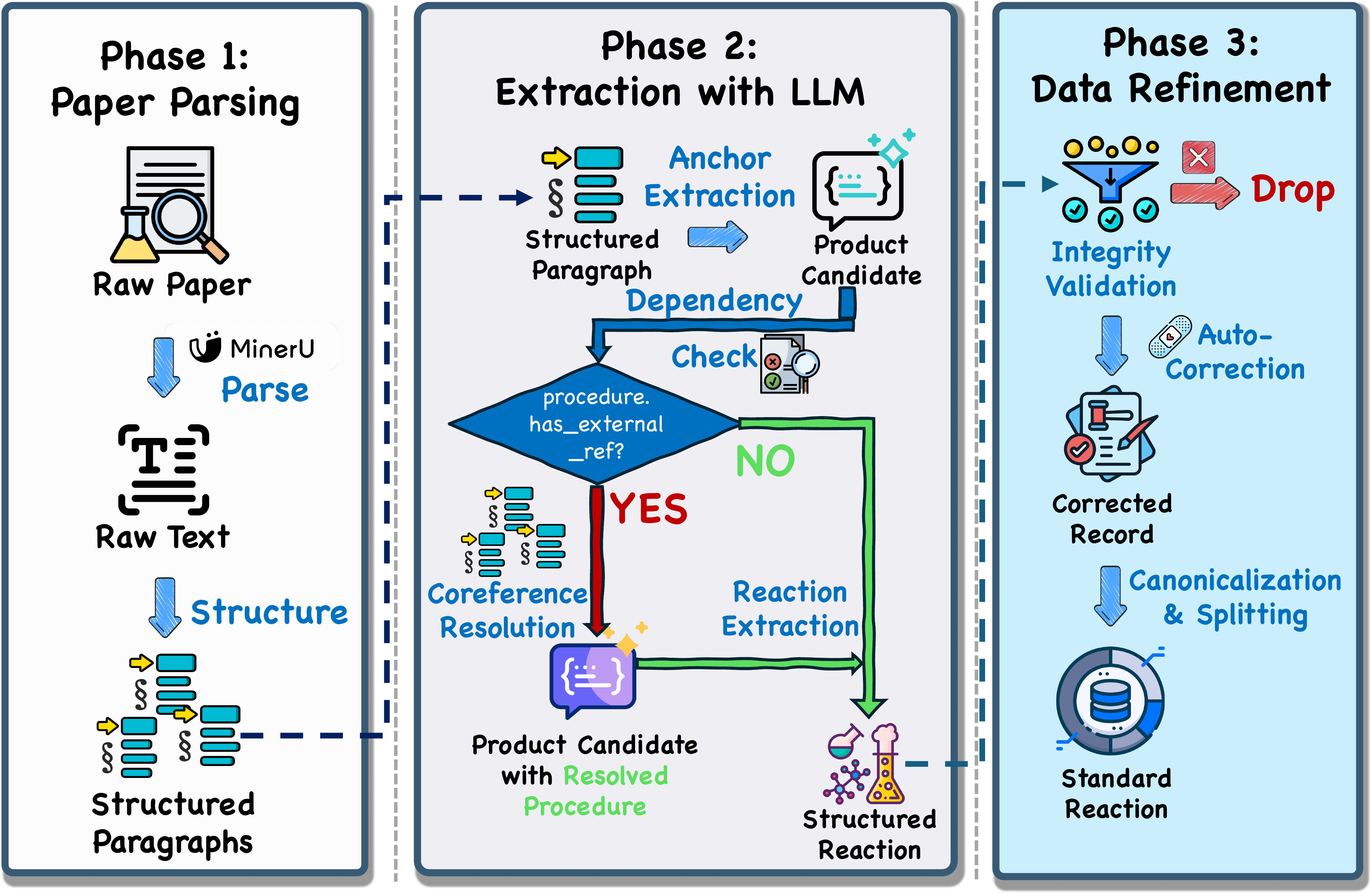}
    \caption{\textbf{The overall architecture of Idt-TE pipeline.} Specifically, the Idt-TE pipeline operates in three stages: Phase 1 digitizes and structures the raw manuscript into logical paragraphs. Phase 2 employs a cascaded, context-aware LLM framework to resolve cross-paragraph dependencies and extract reaction entities based on semantic anchors. Finally, Phase 3 applies rigorous rule-based curation to validate and canonicalize the extracted records, ultimately enabling precision refinement and contextual enrichment of the visual predictions.}
    \label{fig:Idt-TE-pipeline}
\end{figure}
\subsection{Phase 1: Document Parsing and Structuring}
We utilize PDF parsing tools (e.g., MinerU~\cite{wang2024mineru}) to strip layout artifacts from the raw paper, converting it into sequential raw text. The text is subsequently logically chunked into structured paragraphs, providing a clean and bounded semantic context for downstream extraction.

\subsection{Phase 2: Context-Aware LLM Extraction}
Unlike standard sentence-level extraction, chemical procedures often span multiple paragraphs with complex implicit references. To tackle this, we design a cascaded 4-step Large Language Model (LLM) extraction framework.

\vspace{1em}
\noindent\textbf{Step 1: Anchor Extraction.} 
The pipeline adopts a product-centric strategy. The LLM first scans the structured paragraphs to identify and extract all potential product entities as semantic anchors. The specific prompt and examples are detailed in \cref{tab:prompt_anchor}.

\vspace{1em}
\noindent\textbf{Step 2: Dependency Check.} 
For each extracted anchor, the LLM evaluates whether its procedural description is self-contained or relies on external textual references. We categorize these external dependencies into five distinct types, as summarized in \cref{tab:dependency_types}. The prompt for dependency categorization is provided in \cref{tab:prompt_dependency}.

\begin{table}[h]
\centering
\caption{Taxonomy of external procedural dependencies evaluated in Step 2.}
\label{tab:dependency_types}
\resizebox{\linewidth}{!}{
\begin{tabular}{@{}llp{6cm}@{}}
\toprule
\textbf{Type} & \textbf{Semantic Meaning} & \textbf{Example Format} \\
\midrule
\texttt{"title"} & References a specific section title (e.g., Example, Preparation, Intermediate). & \texttt{["Example 3", "title"]} \\
\texttt{"id"} & References a specific compound identifier. & \texttt{["Compound 8", "id"]} \\
\texttt{"iupac\_name"} & References a prior synthesis step via its IUPAC name. & \texttt{["2-chloroaniline", "iupac\_name"]} \\
\texttt{"method"} & References a universally defined general procedure or method. & \texttt{["General Procedure I", "method"]} \\
\texttt{"last\_compound"} & Implicitly inherits the synthesis method of the immediately preceding reaction. & \texttt{["\_\_last\_compound\_\_", "last\_compound"]} \\
\bottomrule
\end{tabular}
}
\end{table}

\vspace{1em}
\noindent\textbf{Step 3: Coreference Resolution.} 
If an external dependency is detected in Step 2, a coreference resolution module is triggered. Guided by the prompt in \cref{tab:prompt_coref}, the LLM retrieves the target text from other paragraphs (e.g., fetching ``General Procedure I'') and rewrites the current procedure by seamlessly integrating the resolved context. This transforms fragmented texts into a fully enriched, self-contained procedure.

\vspace{1em}
\noindent\textbf{Step 4: Reaction Extraction.} 
Operating exclusively on the enriched and self-contained procedures, the LLM executes the final extraction of all chemical entities (reactants, products, catalysts, conditions). The prompt (\cref{tab:prompt_reaction_extraction}) strictly enforces that native identifiers are preserved as primary keys, ensuring direct structural alignment with the visual parsing outputs.

\subsection{Phase 3: Data Refinement}
To mitigate LLM hallucination and format inconsistency, the extracted JSON records are subjected to a rigorous rule-based refinement stage. As illustrated in the final phase of our pipeline, this process operates as a three-step funnel to yield highly standardized datasets:

\vspace{1em}
\noindent\textbf{Step 1: Integrity Validation.} 
This initial gate acts as a strict structural filter. Non-compliant records that fail to meet fundamental constraints are immediately \textbf{dropped}. A record passes this stage only if it satisfies the following criteria:
\begin{itemize}[label=\tiny$\bullet$, nosep]
    \item The \texttt{procedure} field must exist, and its source \texttt{paragraph} must contain $\ge 3$ words.
    \item The \texttt{substances}, \texttt{reactants}, and \texttt{products} arrays must be non-empty, and at least one valid identifier (\texttt{id} or \texttt{iupac\_name}) must be explicitly extracted.
    \item Within the \texttt{substances} list, \texttt{idx} values must increment continuously starting from 0. Each item must exclusively contain the mandatory fields (\texttt{idx}, \texttt{content}, \texttt{chemical\_name}, \texttt{is\_identifier}, \texttt{role}) without extraneous keys, and the \texttt{role} must belong to the permitted closed set.
    \item Product chemical yields must not exceed 100\%.
\end{itemize}

\vspace{1em}
\noindent\textbf{Step 2: Auto-Correction.} 
Records surviving the initial validation undergo automated repairs to resolve logical inconsistencies, yielding a \textbf{Corrected Record}.
\begin{itemize}[label=\tiny$\bullet$, nosep]
    \item \textit{Missing Value Imputation:} Empty \texttt{chemical\_name} fields are automatically filled using their corresponding \texttt{content} strings.
    \item \textit{Role Alignment:} This step enforces strict logical consistency across arrays. It guarantees mutual exclusivity (a single \texttt{idx} cannot appear in multiple role arrays simultaneously) and completeness (every \texttt{idx} must be assigned to at least one role). Any mismatch between a substance's \texttt{role} attribute and its actual array placement is forcefully corrected.
\end{itemize}

\vspace{1em}
\noindent\textbf{Step 3: Canonicalization \& Splitting.} 
The final node formats the corrected records into the ultimate \textbf{Standard Reaction}.
\begin{itemize}[label=\tiny$\bullet$, nosep]
    \item \textit{Textual Normalization:} Typographical artifacts (e.g., non-standard quotes, superfluous hyphens, and spaces) are sanitized. Carbon isotope notations are standardized (e.g., \texttt{13C-labeled} $\rightarrow$ \texttt{[13C]-labeled}).
    \item \textit{Reaction Fission:} Crucially, if an IUPAC name or entity description contains the conjunction ``and'' (representing parallel or mixed synthesis), the record is algorithmically split into multiple independent reaction pathways.
\end{itemize}


\input{Tables/Idt-TE-product-extract-prompt}

\input{Tables/Idt-TE-dependency}

\input{Tables/Idt-TE-paragraph-prompt}

\input{Tables/Idt-TE-reaction-extract-prompt}

%% file: Tables/Idt-TE-product-extract-prompt.tex
\newpage
\vspace{1em}
\begin{center}
    \captionof{table}{Prompt used for Step 1: Anchor Extraction.}
    \label{tab:prompt_anchor}
    
    \begin{tcolorbox}[
        enhanced, breakable, colback=white, colframe=black!75, fonttitle=\bfseries,
        title=Prompt used for Anchor Extraction, boxrule=1pt, arc=0pt, outer arc=0pt,
        left=1ex, right=1ex, top=1ex, bottom=1ex
    ]
    \scriptsize \sffamily \linespread{0.9}\selectfont
    
    You are an expert chemical data extractor. Your task is to identify and extract chemical reactions from the provided experimental text into a structured JSON format.

    \vspace{0.6em}
    \noindent\textbf{1. CORE EXTRACTION CONDITIONS}
    \begin{itemize}[label=$\circ$, nosep, leftmargin=1.5em]
        \item \textbf{Must Extract:} Any step containing a \texttt{procedure} AND at least one product identifier (\texttt{id} OR \texttt{iupac\_name}).
        \item \textbf{Intermediate Products:} MUST be extracted as separate reactions (e.g., "to give compound 3" $\rightarrow$ extract with \texttt{id: "3"}, \texttt{iupac\_name: ""}).
        \item \textbf{Product Location:} Reactants appear at the start of a procedure; products appear at the END (e.g., "...to afford compound 2"). ONLY extract the product's name/ID.
    \end{itemize}

    \vspace{0.6em}
    \noindent\textbf{2. ENTITY RESOLUTION \& CLEANING}
    \begin{itemize}[label=$\circ$, nosep, leftmargin=1.5em]
        \item \textbf{ID Extraction:} Preserve exact alphanumeric IDs (e.g., "3a", "SI-3"). Do not extract generic descriptors ("compound", "product") as IUPAC names.
        \item \textbf{IUPAC Canonicalization:} Remove generic descriptors (e.g., "mixture of", "crude"). Strip all formula-related markup (e.g., \texttt{\{sup X\}}). Do NOT include trailing amount/ID parentheses in the \texttt{iupac\_name}.
        \item \textbf{[CRITICAL] Coreference Resolution:} You MUST replace all relative textual references (e.g., "product of step X", "the above intermediate", "the ester") in the \texttt{procedure} text with their explicitly resolved IUPAC names from previous steps. Do not output unresolved generic references.
    \end{itemize}

    \vspace{0.6em}
    \noindent\textbf{3. MULTI-STEP \& PARALLEL REACTIONS}
    \begin{itemize}[label=$\circ$, nosep, leftmargin=1.5em]
        \item \textbf{Multi-Step Detection:} If a single paragraph isolates multiple different products sequentially, split them into multiple sequential JSON entries.
        \item \textbf{Implicit Reactants:} If a step relies on a previous product but omits it (e.g., "Synthesis of X was similar to Example 1"), append \texttt{" using [previous step's IUPAC product]"} to the \texttt{procedure}.
        \item \textbf{Parallel Reactions:} If X, Y, Z are independently treated to give A, B, C, split into 3 distinct reaction entries immediately.
    \end{itemize}

    \vspace{0.6em}
    \noindent\textbf{4. TRACKING \& IDENTIFIERS (\texttt{title} \& \texttt{step\_id})}
    \begin{itemize}[label=$\circ$, nosep, leftmargin=1.5em]
        \item The (\texttt{title}, \texttt{step\_id}) pair MUST be globally unique.
        \item \textbf{Sequential Chain:} Use the same \texttt{title} and sequential \texttt{step\_id}s (1, 2, 3...) for products in a continuous synthesis chain.
        \item \textbf{Independent Paths:} Use the sub-heading in the \texttt{title} (e.g., "Example 1: Prep of 3a") and \texttt{step\_id: "1"} for independent reactions.
    \end{itemize}

    \vspace{0.6em}
    \noindent\textbf{OUTPUT FORMAT:}
    Return ONLY a JSON array with the following schema:
    \begin{lstlisting}[basicstyle=\ttfamily\scriptsize, breaklines=true, columns=fullflexible]
[
  {
    "title": "Synthesis path title (e.g., 'Example 3: Intermediate 5')",
    "id": "Target compound identifier (e.g., '3a') or ''",
    "iupac_name": "Actual chemical name from procedure end, or ''",
    "mol_num": "Extracted <num> tag value or ''",
    "procedure": "Resolved synthesis procedure text (all references replaced)",
    "step_id": "Step identifier (e.g., '1', '2', '1-A')",
    "needs_keyword_extraction": true if procedure contains unresolved external references (e.g., 'prepared according to Example X'), else false
  }
]
    \end{lstlisting}

    \vspace{0.6em}
    \hrule
    \vspace{0.6em}
    
    \noindent\textbf{Sample Input:}
    \begin{lstlisting}[basicstyle=\ttfamily\scriptsize, breaklines=true, columns=fullflexible]
    
{
  "title": "Example 50",
  "content": [
    {"content": "Compound 50 was prepared similarly to Example 3, 
     substituting 4-bromobenzoic acid for 4-chlorobenzoic acid. 
     Yield: 420 mg (85%)."}
  ]
}
    \end{lstlisting}
    
    \noindent\textbf{Sample Output:}
    \begin{lstlisting}[basicstyle=\ttfamily\scriptsize, breaklines=true, columns=fullflexible]
{
  "title": "Example 50",
  "id": "50",
  "iupac_name": "",             
  "procedure": "Compound 50 was prepared similarly to Example 3, 
   substituting 4-bromobenzoic acid for 4-chlorobenzoic acid. 
   Yield: 420 mg (85%).",        
  "step_id": "1",
  "needs_keyword_extraction": true   
}
    \end{lstlisting}
    
    \end{tcolorbox}
\end{center}
\vspace{1em}

%% file: Tables/Idt-TE-dependency.tex
\vspace{1em}
\begin{center}
    \captionof{table}{Prompt used for Step 2: Dependency Check.}
    \label{tab:prompt_dependency}
    
    \begin{tcolorbox}[
        enhanced, breakable, colback=white, colframe=black!75, fonttitle=\bfseries,
        title=Prompt used for Dependency Check, boxrule=1pt, arc=0pt, outer arc=0pt,
        left=1ex, right=1ex, top=1ex, bottom=1ex
    ]
    \scriptsize \sffamily \linespread{0.9}\selectfont
    
    You are an expert chemical data extractor performing the secondary stage of a pipeline. Your task is to apply strict rules to determine if a chemical procedure explicitly requires external context (keywords) to be fully understood. Keyword extraction is rare; prioritize precision and omit false positives.

    \vspace{0.6em}
    \noindent\textbf{1. EXTRACTION TRIGGERS (When to Extract)}
    \begin{itemize}[label=$\circ$, nosep, leftmargin=1.5em]
        \item \textbf{Explicit Method References:} The procedure uses verbs indicating reliance on another protocol (e.g., "prepared according to Example 3", "analogous to General Procedure A").
        \item \textbf{Ambiguous Compound References:} The procedure references a compound strictly by its origin without providing its IUPAC name or explicit ID (e.g., "the compound of Example 6", "product of Method E").
        \item \textbf{Implicit Prior Reaction:} Short procedures stating "Similarly prepared from [IUPAC name]" without naming a specific Example/Method. Extract as \texttt{[["\_\_last\_compound\_\_", "last\_compound"]]}.
        \item \textbf{Unresolved Self-References:} If the procedure text contains unresolved internal steps (e.g., "the product of step a)", "the piperidine a)") where actual names were not substituted, extract the entry's \textbf{OWN title} (e.g., \texttt{Example 28}) so the fusion stage can retrieve full multi-step context.
    \end{itemize}

    \vspace{0.6em}
    \noindent\textbf{2. NEGATIVE CONSTRAINTS (When NOT to Extract)}
    \begin{itemize}[label=$\circ$, nosep, leftmargin=1.5em]
        \item \textbf{Source Annotations:} Do NOT extract if the reference is merely a parenthetical source annotation accompanying a provided IUPAC name or Compound ID (e.g., "benzyl alcohol (Example 2)", "Compound 3a (from Example 1)").
        \item \textbf{Internal Step Annotations:} Ignore purely parenthetical step markers like "(step 2)" unless they are unresolved textual references as defined above.
        \item \textbf{Self-Contained Procedures:} If the procedure is complete and unambiguous, extract NO keywords.
    \end{itemize}

    \vspace{0.6em}
    \noindent\textbf{3. KEYWORD CANONICALIZATION \& TYPING}
    \\Each extracted keyword must be a tuple: \texttt{[keyword\_string, keyword\_type]}.
    \begin{itemize}[label=$\circ$, nosep, leftmargin=1.5em]
        \item \textbf{Types Definition:} 
        \begin{itemize}[label=--, nosep, leftmargin=1em]
            \item \texttt{"title"}: Examples, Preparations, Intermediates (e.g., "Example 1", "Intermediate 5").
            \item \texttt{"method"}: General reusable protocols (e.g., "General Procedure I", "Method A").
            \item \texttt{"id"}: Pure alphanumeric identifiers or "Compound X".
            \item \texttt{"iupac\_name"}: Explicit chemical names used as procedure pointers.
        \end{itemize}
        \item \textbf{Step Marker Erasure:} Strip all sub-step markers from the keyword string (e.g., "Example 6 Step 1" $\rightarrow$ extract ONLY \texttt{"Example 6"}).
        \item \textbf{Range Expansion:} You MUST expand grouped references (e.g., "Examples 10-12" $\rightarrow$ extract \texttt{"Example 10"}, \texttt{"Example 11"}, and \texttt{"Example 12"} individually).
        \item \textbf{IUPAC Cleaning:} Strip descriptive prefixes from IUPAC references (e.g., "the preparation of benzene" $\rightarrow$ extract ONLY \texttt{"benzene"}).
    \end{itemize}

    \vspace{0.6em}
    \noindent\textbf{OUTPUT FORMAT:}
    Return the input JSON object with an appended \texttt{keyword} array. If NO valid keywords are found based on the strict criteria, omit the \texttt{keyword} field entirely.

    \begin{lstlisting}[basicstyle=\ttfamily\scriptsize, breaklines=true, columns=fullflexible]
{
  "title": "Example 5",
  "id": "5",
  "iupac_name": "benzyl chloride",
  "procedure": "Prepared according to Example 3, using the compound from Method A.",
  "keyword": [["Example 3", "title"], ["Method A", "method"]]
}
    \end{lstlisting}

    \vspace{0.6em}
    \hrule
    \vspace{0.6em}
    
    \noindent\textbf{Sample Input:}
    \begin{lstlisting}[basicstyle=\ttfamily\scriptsize, breaklines=true, columns=fullflexible]
{
  "title": "Example 50",
  "id": "50",
  "iupac_name": "",             
  "procedure": "Compound 50 was prepared similarly to Example 3, 
   substituting 4-bromobenzoic acid for 4-chlorobenzoic acid. 
   Yield: 420 mg (85%).",        
  "step_id": "1",
  "needs_keyword_extraction": true   
}
    \end{lstlisting}
    
    \noindent\textbf{Sample Output:}
    \begin{lstlisting}[basicstyle=\ttfamily\scriptsize, breaklines=true, columns=fullflexible]
{
  "title": "Example 50",
  "id": "50",
  "iupac_name": "",             
  "procedure": "Compound 50 was prepared similarly to Example 3, 
   substituting 4-bromobenzoic acid for 4-chlorobenzoic acid. 
   Yield: 420 mg (85%).",        
  "step_id": "1",
  "needs_keyword_extraction": true,
  "keyword": [["Example 3", "title"]]
}
    \end{lstlisting}
    
    \end{tcolorbox}
\end{center}
\vspace{1em}

%% file: Tables/Idt-TE-paragraph-prompt.tex
\newpage
\vspace{1em}
\begin{center}
    \captionof{table}{Prompt used for Step 3: Coreference Resolution.}
    \label{tab:prompt_coref}
    
    \begin{tcolorbox}[
        enhanced, breakable, colback=white, colframe=black!75, fonttitle=\bfseries,
        title=Prompt used for Coreference Resolution, boxrule=1pt, arc=0pt, outer arc=0pt,
        left=1ex, right=1ex, top=1ex, bottom=1ex
    ]
    \scriptsize \sffamily \linespread{0.9}\selectfont
    
    You are an expert chemist. Your task is to resolve cross-paragraph coreferences by intelligently fusing a target fragmented procedure with its corresponding referential text, producing a complete, self-contained experimental procedure.

    \vspace{0.6em}
    \noindent\textbf{1. CORE MISSION \& INPUT STRUCTURE}
    \begin{itemize}[label=$\circ$, nosep, leftmargin=1.5em]
        \item \textbf{Input:} You will receive a \texttt{TARGET MOLECULE} block (containing vague references like "as in Example 1") and a \texttt{REFERENTIAL INFORMATION} block (containing the full text of referenced Examples/Methods).
        \item \textbf{Action:} REPLACE all textual reference patterns with actual experimental details extracted from the reference text.
        \item \textbf{Output:} A fully self-contained procedure with \textbf{ZERO} references to external methods.
    \end{itemize}

    \vspace{0.6em}
    \noindent\textbf{2. RESOLUTION STRATEGIES}
    \begin{itemize}[label=$\circ$, nosep, leftmargin=1.5em]
        \item \textbf{Explicit Substitution ("except A was substituted for B"):} Extract the COMPLETE reference procedure. Replace every occurrence of B with A. Preserve all other conditions.
        \item \textbf{Implicit Inference (No explicit substitution):} If the target says "similar to Example X", you MUST infer the correct starting material based on the TARGET's IUPAC product structure via retrosynthetic logic. \textbf{DO NOT} blindly copy the reference's starting material, or you will synthesize the wrong molecule.
        \item \textbf{Multi-Step Alignment:} If the reference contains a multi-step sequence (e.g., Steps A to G), extract ONLY the step(s) structurally relevant to the target molecule. If the target explicitly names a step (e.g., "method of Example 1, step c"), you MUST extract exactly that step.
        \item \textbf{Compound Pointer Resolution:} Replace patterns like "the compound of Example X" or "Example XA (amount)" strictly with the actual IUPAC name of Example X's product/reactant. Do not append the entire procedure.
    \end{itemize}

    \vspace{0.6em}
    \noindent\textbf{3. CRITICAL CONSTRAINTS (ZERO TOLERANCE)}
    \begin{itemize}[label=$\circ$, nosep, leftmargin=1.5em]
        \item \textbf{No Residual References:} The output MUST NOT contain phrases like "as described in", "similar to", "following Example", or "the compound of Example X".
        \item \textbf{Resolve Internal Steps:} You MUST replace internal pointers like "the product of step a)" using the provided \texttt{[SELF-REFERENCE]} context block.
        \item \textbf{Preserve Explicit Reactants:} If the target procedure already explicitly names reactants (e.g., "starting from [Compound Z]"), you MUST keep them. Only extract the \textit{reaction conditions} (catalyst, temp, solvent) from the reference.
        \item \textbf{Quantitative Transfer:} ALL reagent amounts (mass, volume, mmol, equiv) from the reference MUST be preserved. \textbf{EXCEPTION:} NEVER copy the product yield/mass from the reference; use only the target's yield if provided.
        \item \textbf{No Generic Placeholders:} Replace vague terms like "appropriate starting material" with specific inferred chemical names.
    \end{itemize}

    \vspace{0.6em}
    \noindent\textbf{OUTPUT FORMAT:}
    Return ONLY a valid JSON object:
    \begin{lstlisting}[basicstyle=\ttfamily\scriptsize, breaklines=true, columns=fullflexible]
{
  "completed_reaction": "The fully resolved, standalone procedure string.",
  "merge_summary": "Brief explanation of the extraction/inference logic applied.",
  "generic_terms_replaced": ["List of generic terms replaced with specific IUPAC names."]
}
    \end{lstlisting}

    \vspace{0.6em}
    \hrule
    \vspace{0.6em}
    
    \noindent\textbf{Sample Input:}
    \begin{lstlisting}[basicstyle=\ttfamily\scriptsize, breaklines=true, columns=fullflexible]
{
  "title": "Example 50",
  "id": "50",
  "iupac_name": "",             
  "procedure": "Compound 50 was prepared similarly to Example 3, 
   substituting 4-bromobenzoic acid for 4-chlorobenzoic acid. 
   Yield: 420 mg (85%).",        
  "step_id": "1",
  "needs_keyword_extraction": true,
  "keyword": [["Example 3", "example"]]
}
    \end{lstlisting}
    
    \noindent\textbf{Sample Output:}
    \begin{lstlisting}[basicstyle=\ttfamily\scriptsize, breaklines=true, columns=fullflexible]
{
  "title": "Example 50",
  "id": "50",
  "iupac_name": "",
  "procedure": "To a solution of 4-bromobenzoic acid (500 mg) 
   in MeOH (10 mL) was added H2SO4 (cat.). The reaction 
   was stirred at RT for 12 h. The mixture was concentrated 
   and purified to give compound 50 (420 mg, 85%).",
  "step_id": "1",
  "needs_keyword_extraction": true
}
    \end{lstlisting}
    
    \end{tcolorbox}
\end{center}
\vspace{1em}

%% file: Tables/Idt-TE-reaction-extract-prompt.tex
\newpage
\vspace{1em}
\begin{center}
    \captionof{table}{Prompt used for Step 4: Reaction Extraction \& Structuring.}
    \label{tab:prompt_reaction_extraction}
    
    \begin{tcolorbox}[
        enhanced, breakable, colback=white, colframe=black!75, fonttitle=\bfseries,
        title=Prompt used for Reaction Extraction \& Structuring, boxrule=1pt, arc=0pt, outer arc=0pt,
        left=1ex, right=1ex, top=1ex, bottom=1ex
    ]
    \scriptsize \sffamily \linespread{0.9}\selectfont
    
    You are an expert chemical data extractor. Your task is to identify key chemical reaction parameters from the provided text and output them into a strict, pointer-based JSON schema. 

    \vspace{0.6em}
    \noindent\textbf{1. ENTITY CLASSIFICATION \& EXCLUSIONS}
    \begin{itemize}[label=$\circ$, nosep, leftmargin=1.5em]
        \item \textbf{Catalysts vs. Reagents:} Metal complexes, ligands (e.g., -Phos), organocatalysts (e.g., DMAP), and chemicals in catalytic amounts ("drops") are \texttt{catalyst}. Consumed/stoichiometric chemicals are \texttt{reagent}.
        \item \textbf{Workup vs. Reaction Steps:} DO NOT extract chemicals used purely for workup (washing, drying, filtration aids like Celite). DO extract chemicals from late-stage reactions (e.g., salt formation, hydrogenation). Workup \textit{text} goes strictly into the \texttt{workup} field.
        \item \textbf{Intermediates:} Do NOT extract transient intermediates generated and consumed within the same contiguous procedure.
    \end{itemize}

    \vspace{0.6em}
    \noindent\textbf{2. ENTITY SPLITTING RULES}
    \begin{itemize}[label=$\circ$, nosep, leftmargin=1.5em]
        \item \textbf{Alternatives:} Split slash-separated alternatives (e.g., "Et3N / pyridine" $\rightarrow$ two entries).
        \item \textbf{Solvates/Adducts:} Split adducts (e.g., "Pd(dppf)Cl2.DCM" $\rightarrow$ catalyst + solvent). \textit{Exception:} Keep hydrates intact (e.g., "NiCl2.6H2O").
        \item \textbf{Solutions:} Split "X in Y" patterns (e.g., "4M HCl in dioxane" $\rightarrow$ solute HCl + solvent dioxane). Assign amounts accordingly.
    \end{itemize}

    \vspace{0.6em}
    \noindent\textbf{3. CANONICALIZATION \& IDENTIFIER RULES}
    \begin{itemize}[label=$\circ$, nosep, leftmargin=1.5em]
        \item \textbf{Amounts \& Units:} Aggregate all quantities (mass, volume, mmol, equiv) into \texttt{amount}. Cap \texttt{yield\_ratio} at 100\%. Standardize temperatures (e.g., \texttt{degC}).
        \item \textbf{Prefix Stripping:} Remove descriptive prefixes from identifiers (e.g., "Compound 1a", "pyrrole 5" $\rightarrow$ extract only \texttt{"1a"}, \texttt{"5"} with \texttt{is\_identifier: true}).
        \item \textbf{ID vs. IUPAC Resolution:} 
        \begin{itemize}[label=--, nosep, leftmargin=1em]
            \item \textit{IUPAC with ID in parens} (e.g., "benzene (1)"): \texttt{content} = benzene, \texttt{chemical\_name} = benzene, \texttt{is\_identifier: false}.
            \item \textit{ID + descriptive prefix} (e.g., "amine 3a"): \texttt{content} = 3a, \texttt{chemical\_name} = amine.
        \end{itemize}
        \item \textbf{SMILES Preparation:} Remove formula markup (e.g., \texttt{\{sup X\}}), stoichiometric ratios, and descriptive terms ("derivatives") from \texttt{chemical\_name}. Fix obvious bracket typos.
    \end{itemize}

    \vspace{0.6em}
    \noindent\textbf{4. UNIFIED SCHEMA \& IDX POINTERS}
    \begin{itemize}[label=$\circ$, nosep, leftmargin=1.5em]
        \item \textbf{Single Source of Truth:} All unique reaction chemicals reside in the \texttt{substances} array.
        \item \textbf{Role Arrays:} The top-level \texttt{reactants}, \texttt{catalyst}, \texttt{reagents}, and \texttt{solvent} fields contain ONLY integer \texttt{idx} pointers to the \texttt{substances} array.
        \item \textbf{Stages:} Group physical conditions (time, temp) into sequential \texttt{stages}. Reference participating chemicals via \texttt{idx} arrays.
    \end{itemize}

    \vspace{0.6em}
    \noindent\textbf{OUTPUT FORMAT:} Return ONLY a valid JSON object matching this schema:
    \begin{lstlisting}[basicstyle=\ttfamily\scriptsize, breaklines=true, columns=fullflexible]
{
  "substances": [
    {
      "idx": 0,
      "content": "Original raw text (e.g., '1a' or 'benzene')",
      "amount": "Aggregated amounts with units (e.g., '100 mg, 0.5 mmol')",
      "chemical_name": "Cleaned name for SMILES conversion",
      "is_identifier": true,
      "equivalence": 0.0,
      "mmol": 0.0,
      "role": "reactant" // Must be: reactant/catalyst/reagent/solvent
    }
  ],
  "reactants": [0], // Array of idx pointers
  "catalyst": [],
  "reagents": [],
  "solvent": [],
  "products": [
    {
      "content": "target compound",
      "production": "e.g., 59 mg",
      "yield_ratio": "e.g., 85%",
      "conversion_rate": "", "stereo_selectivity": "", "ee": "", "dr": "", "rr": "",
      "appearance": "e.g., white solid",
      "chemical_name": "IUPAC product name",
      "is_identifier": false
    }
  ],
  "stages": [
    {
      "stage_id": "stage_1",
      "substances": [0], // idx pointers of substances added in this stage
      "time": "2 h",
      "temperature": "80degC",
      "atmosphere": "N2",
      "pressure": "", "PH": "", "stirring_speed": "", 
      "vacuum_condition": "", "light_condition": "", "cooling_heating_condition": "",
      "workup": "Original text describing extraction, washing, drying..."
    }
  ]
}
    \end{lstlisting}

    \vspace{0.6em}
    \hrule
    \vspace{0.6em}
    
    \noindent\textbf{Sample Input:}
    \begin{lstlisting}[basicstyle=\ttfamily\scriptsize, breaklines=true, columns=fullflexible]
{
  "title": "Example 50",
  "id": "50",
  "iupac_name": "",
  "procedure": "To a solution of 4-bromobenzoic acid (500 mg) 
   in MeOH (10 mL) was added H2SO4 (cat.). The reaction 
   was stirred at RT for 12 h. The mixture was concentrated 
   and purified to give compound 50 (420 mg, 85%).",
  "step_id": "1",
  "needs_keyword_extraction": true
}
    \end{lstlisting}
    
    \noindent\textbf{Sample Output:}
    \begin{lstlisting}[basicstyle=\ttfamily\scriptsize, breaklines=true, columns=fullflexible]
{
  "title": "Example 50",
  "id": "50",
  "iupac_name": "",
  "procedure": {
    "paragraph": "To a solution of 4-bromobenzoic acid (500 mg)...",
    "substances": [
      {"idx": 0, "content": "4-bromobenzoic acid", 
       "amount": "500 mg", "role": "reactant", ...},
      {"idx": 1, "content": "MeOH", 
       "amount": "10 mL", "role": "solvent", ...},
      {"idx": 2, "content": "H2SO4", 
       "amount": "", "role": "catalyst", ...}
    ],
    "reactants": [0],
    "reagents": [],
    "catalyst": [2],
    "solvent": [1],
    "products": [{"content": "50", "is_identifier": true, 
                   "yield_ratio": "85%"}],
    "stages": [{"stage_id": "stage_1", "substances": [0,1,2],
                "temperature": "RT", "time": "12 h"}]
  }
}
    \end{lstlisting}
    
    \end{tcolorbox}
\end{center}
\vspace{1em}